\documentclass{article}
\usepackage{hyphenat}
\usepackage{microtype}
\usepackage{graphicx}
\graphicspath{{figures/}{sec/}} 
\usepackage{subcaption}
\usepackage{booktabs} 
\usepackage{multirow} 
\usepackage{longtable}
\usepackage{float}
\usepackage{caption}

\usepackage{hyperref}
\usepackage[ruled,vlined]{algorithm2e}

\usepackage{wrapfig}\usepackage[table]{xcolor} 
\definecolor{bestcolor}{RGB}{219, 208, 237}
\definecolor{secondcolor}{RGB}{241, 237, 248}
\definecolor{line-blue}{RGB}{243, 248, 252}
\definecolor{taskSpatial}{RGB}{255, 204, 170}   
\definecolor{taskTransform}{RGB}{210, 190, 230} 
\definecolor{taskLogic}{RGB}{160, 200, 255}     
\definecolor{taskAbstract}{RGB}{245, 170, 185}  
\definecolor{taskPercept}{RGB}{180, 225, 200}   
\definecolor{taskPhysics}{RGB}{30, 60, 120}

\definecolor{resourceblue}{HTML}{4C7DBF}

\newcommand{\resource}[2]{\href{#2}{\textcolor{resourceblue}{[\textbf{#1}]}}}

\usepackage{xspace}

\usepackage[pr]{icml2026}

\makeatletter
\gdef\affil@sep{\enspace}
\newcommand{\affilbreak}{\gdef\affil@sep{\\[2pt]}}

\renewcommand{\printAffiliationsAndNotice}[1]{\global\icml@noticeprintedtrue%
  \stepcounter{@affiliationcounter}%
  \noindent\begin{minipage}{\textwidth}%
    \centering
    \fontsize{9}{11}\selectfont%
    \forloop{@affilnum}{1}{\value{@affilnum} < \value{@affiliationcounter}}{%
      \textsuperscript{\arabic{@affilnum}}\,\csname @affilname\the@affilnum\endcsname%
      \affil@sep\gdef\affil@sep{\enspace}%
    }%
    \ificmlshowauthors #1\fi%
    \ifdefined\icmlcorrespondingauthor@text
      \\[4pt]%
      \textsuperscript{\textdagger}\,Correspondence to: \icmlcorrespondingauthor@text.%
    \fi%
    \\[6pt]
    \resource{Website}{https://object-permanence.world}\enspace
    \resource{Data}{https://object-permanence.world/data}\enspace
    \resource{Code}{https://object-permanence.world/code}\enspace
    \resource{Benchmark}{https://object-permanence.world/benchmark}\enspace
    \resource{Model}{https://object-permanence.world/model}\enspace
    \resource{Leaderboard}{https://object-permanence.world/leaderboard}%
  \end{minipage}%
  \vskip 8pt%
  {\centering\textcolor{pr@lightgray}{\rule{0.2\textwidth}{0.3pt}}\par}%
}
\makeatother

\usepackage{amsmath}
\usepackage{amssymb}
\usepackage{mathtools}
\usepackage{amsthm}

\usepackage[capitalize,noabbrev]{cleveref} 

\theoremstyle{plain}

\theoremstyle{definition}

\theoremstyle{remark}

\usepackage[disable,textsize=tiny]{todonotes}
\usepackage{enumitem}

\usepackage{makecell}

\begin{document}
\twocolumn[
  \icmltitle{Training Object Permanence in World Models}



\icmlsetsymbol{equal}{*}

\begin{icmlauthorlist}
    \icmlauthor{Haotian Zhang}{usc,equal}
    \icmlauthor{Fengyuan Yu}{cmu,equal}
    \icmlauthor{Dezhi Luo}{umich,equal}
    \icmlauthor{Haoran Sun}{jhu}
    \icmlauthor{Zehong Zhao}{ucsd}
    \icmlauthor{Qingying Gao}{jhu}
    \icmlauthor{Yihan Li}{cmu}
    \icmlauthor{Siyuan An}{cmu}
    \icmlauthor{Huayi Qin}{cmu}
    \icmlauthor{Yilan Zhang}{ucla}
    \icmlauthor{Zhengze Jiang}{columbia}
    \icmlauthor{Pinyuan Feng}{columbia}
    \icmlauthor{Renrui Zhang}{cmu}
    \icmlauthor{Ziyu Guo}{cmu}
    \icmlauthor{Letian Wang}{toronto}
    \icmlauthor{Mengyue Yang}{bristol}
    \icmlauthor{Kangfu Mei}{jhu}
    \icmlauthor{Maijunxian Wang}{ucb}
    \icmlauthor{Ran Ji}{ucsd}
    \icmlauthor{Vikash Kumar}{columbia}
    \icmlauthor{Freda Shi}{waterloo}
    \icmlauthor{Chandra Sripada}{umich}
    \icmlauthor{Vincent C. Müller}{fau} 
    \icmlauthor{Philip Torr}{oxford}
    \icmlauthor{Alan Yuille}{jhu}
    \icmlauthor{Nikolaus Kriegeskorte}{columbia}
    \icmlauthor{Felix Juefei-Xu}{nyu}
    \icmlauthor{Lvmin Zhang}{stanford}
    \icmlauthor{Jieneng Chen}{stanford}
    \icmlauthor{Yilun Du}{harvard}
    \icmlauthor{Hokin Deng}{cmu}
\end{icmlauthorlist}

\icmlaffiliation{toronto}{University of Toronto}
\icmlaffiliation{bristol}{University of Bristol} 
\icmlaffiliation{usc}{University of Southern California}
\icmlaffiliation{waterloo}{University of Waterloo}
\icmlaffiliation{fau}{Friedrich-Alexander-Universität Erlangen}
\icmlaffiliation{nyu}{New York University}
\icmlaffiliation{umich}{University of Michigan}
\icmlaffiliation{ucsd}{University of California, San Diego}
\icmlaffiliation{ucla}{University of California, Los Angeles}
\icmlaffiliation{ucb}{University of California, Berkeley}
\icmlaffiliation{jhu}{Johns Hopkins University}
\icmlaffiliation{cmu}{Carnegie Mellon University}
\icmlaffiliation{columbia}{Columbia University}
\icmlaffiliation{harvard}{Harvard University}
\icmlaffiliation{oxford}{University of Oxford}
\icmlaffiliation{stanford}{Stanford University}

\icmlcorrespondingauthor{Hokin Deng}{hokind@andrew.cmu.edu}
\printAffiliationsAndNotice{\icmlEqualContribution}

\icmlkeywords{Object Permanence, Video Models, World Models, Core Knowledge}
]




\begin{figure*}[h]\centering\makebox[\textwidth][c]{\includegraphics[width=1\textwidth]{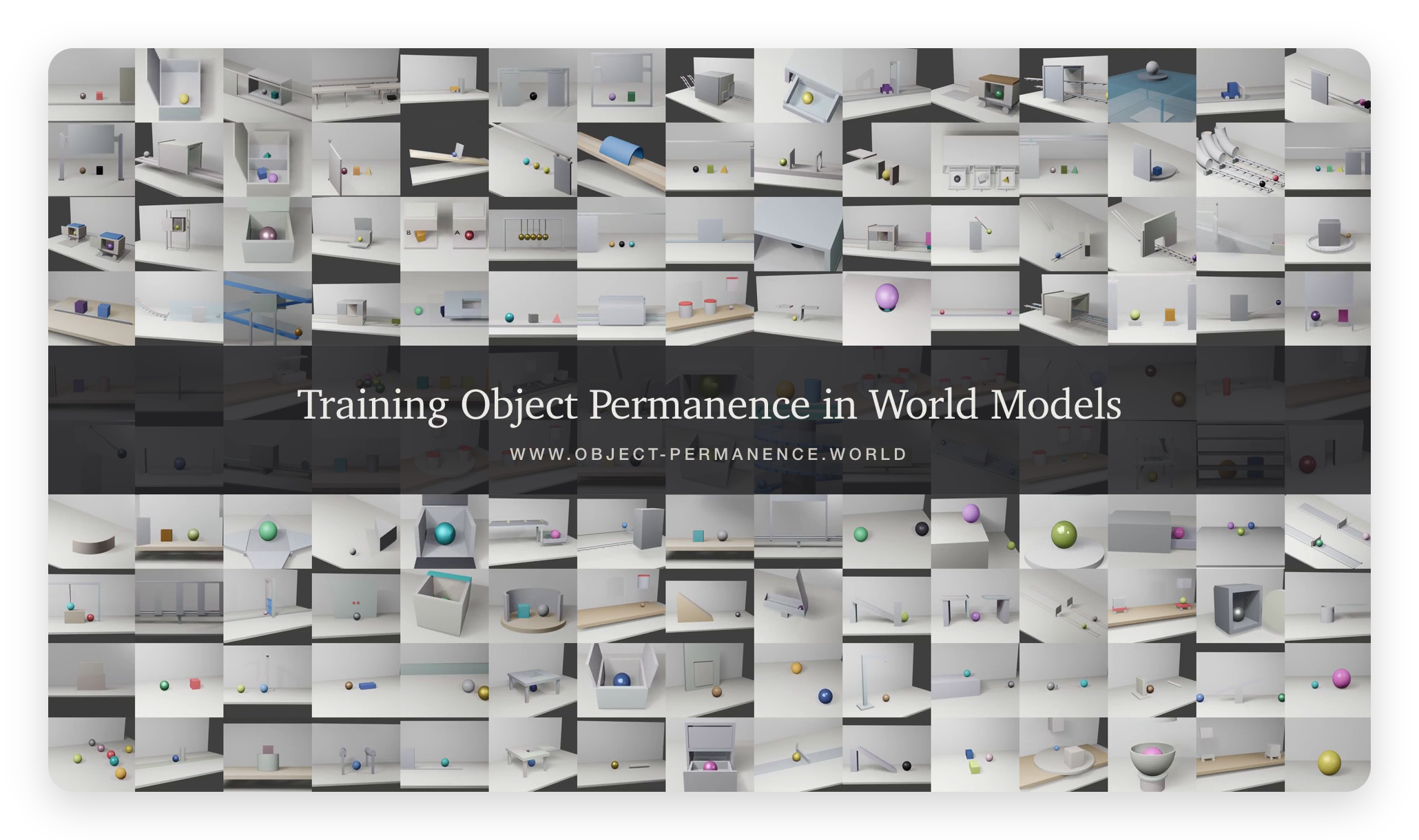}} \caption{\textbf{WROP: World Reasoning with Object Permanence.} We introduce a benchmark and training resource for object permanence and solidity in video generation models, built on 150 hand-designed Blender generators across six task families. Each generator randomises lighting, camera angle, speed, and other nuisance parameters while preserving the task's core cognitive-scientific structure, yielding 10{,}000 samples per task. We release this 1.5M-sample training corpus, a 300-question evaluation exam with human Elo ratings across 14 video generation models, and PWM-WROP, a fine-tuned continuation model trained on WROP data achieving state-of-the-art performance in its category.}
\end{figure*}

\begin{abstract}
Object permanence and solidity are hallmarks of human cognitive priors. Recent studies show that video generation models, a paradigmatic class of current world models, have begun to show emerged reasoning abilities, making them ideal candidates for building human-like physical intelligence. Do video models have emerged object permanence in them? If not, could we train them with a core-cognition inspired dataset? We introduce WROP (World Reasoning with Object Permanence), a data infrastructure of 150 hand-designed cognitive science inspired tasks, divided into six cognitive categories. We build Blender generators that randomize speed, lighting, camera angle, and other nuisance parameters while preserving each task's cognitive structure, yielding 10,000+ samples per task. We release a 1.5M-sample training corpus and a 300-question exam. On this exam we evaluate 14 video models: 3 reference-to-video, 7 edit, and 4 continuation, among which PWM-WROP, our 16B world model. In a blind pairwise Elo study, PWM-WROP ranks first among continuation models and third overall, behind only a statistical tie between two reference-to-video models. We release the data, exam, model answers, scores, weights, and PWM, our native-PyTorch training stack on AWS Trainium2.

\end{abstract}
\section{Introduction}
Recent video generation models produce photorealistic, temporally coherent footage, and on this basis they are increasingly regarded as world models capable of simulating the world \citep{openai2024sora, ho2020denois,deepmind2025veo3, kuaishou2025kling26, kong2024hunyuan, wan2025wan, peebles2023scala, nvidia2026cosmos3, wang2026very, xu2026vbvrpro, nvidia2025cosmosworldfoundationmodel}. Yet a characteristic failure persists: objects vanish behind occluders and re-emerge at impossible positions, or pass through solid barriers undeflected. These failures concern foundational aspects of physical intelligence in humans: \textbf{object permanence (OP)} and \textbf{object solidity (OS)}. Infants represent occluded objects by 3.5 months \citep{baillargeon1986representing,stahl2015observing} and register solidity violations within the first half-year \citep{baillargeon1985object,hespos2001reasoning}. Both OP and OS are considered to be part of core knowledge \citep{spelke2007core}: domain-specific representational systems that are operational early in development and provide the scaffolding for subsequent physical inference \citep{carey2009origin}. Likewise, OP and OS failures in video generation could be structurally upstream: a model that permits interpenetration cannot produce physically valid collision or support-removal events, and any higher-level scene construction or causal reasoning is likely to inherit these errors. As such, evaluating, understanding, and enabling OP and OS in video generation models is an important open challenge. 
 
We introduce WROP (World Reasoning with Object Permanence), a dedicated 3D synthetic benchmark for video reasoning constructed in Blender. WROP comprises 150 self-contained Blender generators organized across six cognitively grounded task families (three probing OP, three OS), released with a 1.5-million-sample training corpus (10,000 samples per generator) and a fixed 300-question exam (two questions per generator). Each ground-truth clip is split around its key physical event: models receive the input half as context and are tasked with generating the target half, which contains the event and its consequence. Generated continuations are judged by human evaluators against physically consistent, hand-authored animation ground truth, circumventing the core knowledge limitations that disqualify VLM judges for this setting \citep{li2025core, luo2025philosophical, luo2025machine, luo2026vision}. We also ask whether OP and OS can be trained: we have post-trained PWM-WROP, a 16B world model, on our dataset.

Across 14 models spanning three interface classes (4 continuation, 3 reference-to-video, and 7 edit models), a blind pairwise study of 20 raters (Bradley--Terry ratings on the Elo scale with rater-clustered bootstrap intervals) places PWM-WROP first among continuation models at Elo 1679.5, behind a statistical tie between two commercial reference-to-video systems at 1723.6 (Section~\ref{sec:results}, Figure~\ref{fig:elo}). This ranking is achieved at a native output resolution of 320$\times$192, compared to 720p and 1080p outputs from competing systems; at matched resolution, PWM-WROP obtains the best LPIPS and MS-SSIM against the target video.

In summary, WROP establishes a principled foundation for evaluating and training object permanence and solidity in video generation models. It provides: 1) a cognitively grounded benchmark and training corpus built on hand-authored generators spanning six object-permanence and solidity task families, with per-sample trajectories, scene-state metadata, and a fixed evaluation exam; 2) a fine-tuned continuation model trained on this corpus, ranking first among true-continuation models in a blind pairwise human study competitive with frontier commercial systems; 3) a comprehensive evaluation framework combining human preference judgments with automatic target-fit metrics, coming with detailed qualitative analyses across representative tasks; 4) a native-PyTorch training stack with the full engineering record for reproducing and extending the model. Together, these components make physical reasoning in video generation trainable on cognitively principled data, evaluable with human-grounded criteria, and experimentally controllable through structured task design, which we consider a critical step in building world models with human-like physical reasoning capabilities.


\section{Related Works}

\paragraph{Video Models as World Models.} The modern video generation landscape emerged from the introduction of denoising diffusion probabilistic models \citep{ho2020denois} and their subsequent scaling through transformer-based architectures \citep{peebles2023scala,blattmann2023align}. Frontier proprietary systems including Sora \citep{openai2024sora}, MovieGen \citep{polyak2024movie}, Veo 3.1 \citep{deepmind2025veo3}, and Kling 2.6 \citep{kuaishou2025kling26}, have demonstrated impressive perceptual fidelity and temporal coherence; open-source counterparts, including Wan2.2 \citep{wan2025wan}, HunyuanVideo \citep{kong2024hunyuan}, CogVideoX-1.5 \citep{yang2024cogvid}, and LTX-2 \citep{hacohen2026ltx2}, have achieved comparable capabilities. A growing body of work probes reasoning capabilities in these models \citep{wang2026very,guo2025mmecof,liu2025genvire,cai2025mmgr,wiedemer2025videozeroshot,yang2025vrbench,he2025rulerbench}, demonstrating promising performance on tasks such as maze-solving, temporal induction, and abstract rule following, and establishing video generation models as increasingly plausible candidates for the role of world models capable of simulating structured physical environments \citep{lecun2022path}. Despite these advances, existing investigations share notable gaps in their targets: (1) coverage is predominantly limited to two-dimensional environments; (2) evaluations are limited to image-to-video generation and do not cover the video-to-video (V2V) setting, an important locus of inference with substantial real-world use cases; and (3) no benchmark provides dedicated evaluation of structured physical inference about object identity, physical constraints, and causal consequences that are foundational to human-like world models. Existing benchmarks also share structural limitations: small aggregate scale per task, absent or minimal training splits, and prevalent reliance on VLM-based scoring \citep{xu2026vbvrpro}. The last limitation is particularly consequential for benchmarks targeting intuitive physics: multimodal language models exhibit systematic core knowledge deficits \citep{li2025core}, fail to reason about physical transformation \citep{luo2026vision}, and lack reliable perceptual constancy \citep{sun2025probing}, rendering them unreliable judges for precisely the capacities under test. VR-OP\&S addresses all of these gaps by following a core-cognition approach: a large training data repository based on strictly operationalized cognitive-scientific task paradigms that enables native evaluation of object permanence and solidity in three-dimensional environments.

\paragraph{Object Permanence and Solidity: Cognitive and Philosophical Foundations.} The principle that objects persist through time and space independently of observation has roots in both philosophy and developmental science. \citet{kant1929critique} identified the continued existence of objects as a formal precondition of experience, a view that resonates with \citet{wittgenstein1976cause}, who argued that causal intuition is grounded in primitive perceptual awareness rather than learned inference. \citet{piaget1954construction} treated object permanence as the defining cognitive achievement of the sensorimotor stage, proposing that it develops gradually through action-based experience. Subsequent experimental work further refined this account: violation-of-expectation (VoE) paradigms established that infants represent the continued existence and location of occluded objects from as early as 3.5 months \citep{baillargeon1986representing}, form expectations about containment well before the end of the first year \citep{hespos2001reasoning}, and respond to unexpected violations with measurable orienting and exploratory behaviour \citep{stahl2015observing,bremner2015perception}. Object solidity emerges with comparable precocity \citep{sanford1967volume}: infants distinguish between events that respect and violate the impenetrability of solid surfaces within the first half-year of life \citep{baillargeon1985object,hespos2012physics}, extend this constraint to animate agents \citep{saxe2006five}, and use it to predict the outcomes of support-removal events \citep{hood2000predicting}. \citet{falck2020core} further demonstrate that solidity constraints persist as automatic, non-inferential responses in adult visual cognition, even when they dissociate from explicit reasoning. Together, this body of work establishes OP and OS as the most primitive layer of the core knowledge system \citep{spelke2007core}: constitutive features of physical intelligence that any general physical reasoning system ought to instantiate \citep{carey2011precis,long2024nativism,luo2025philosophical}.
\vspace{-2mm}

\section{Dataset}
 
We describe the cognitive taxonomy underlying our task design (Section~\ref{sec:taxonomy}), present key dataset statistics and the release contents (Section~\ref{sec:stats}), and detail the data generation pipeline (Section~\ref{sec:pipeline}).

\vspace{-2mm}

\begin{figure*}[h]\centering\makebox[\textwidth][c]{\includegraphics[width=1.0\textwidth]{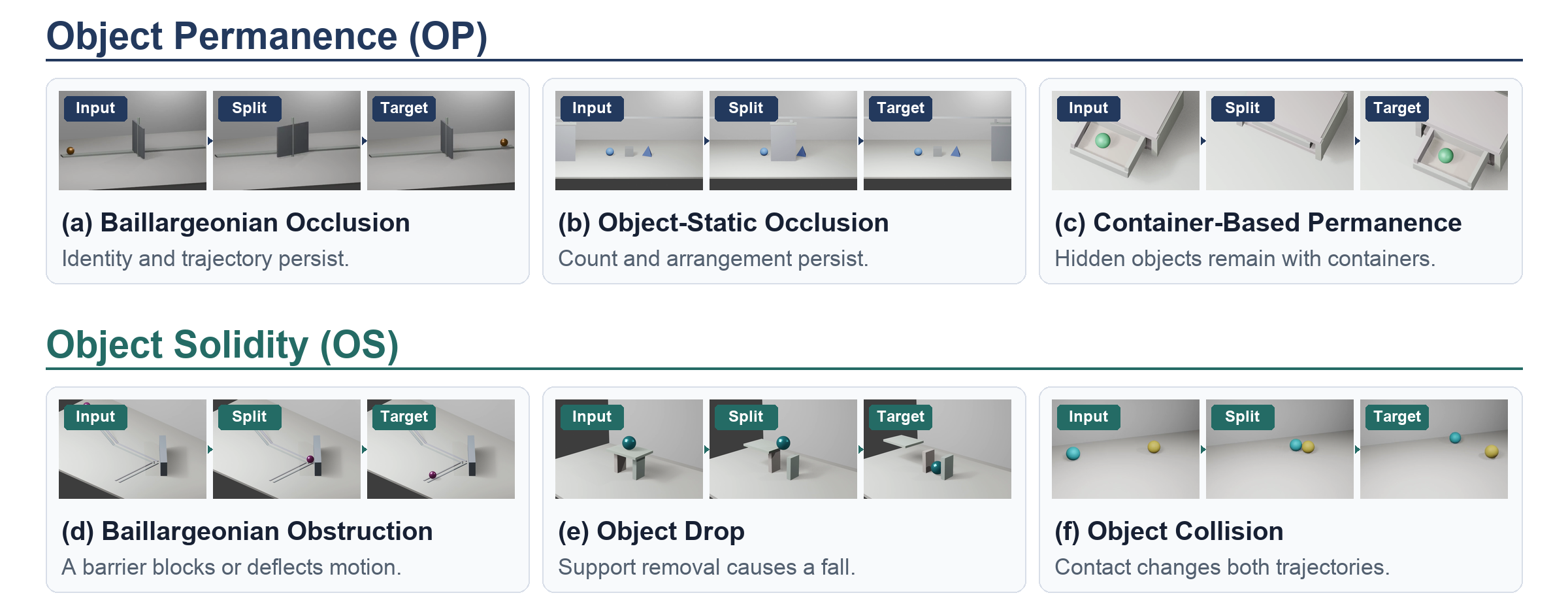}}\vspace{6pt} \caption{Taxonomy of the six WROP task families. The top row evaluates object permanence through dynamic occlusion, static-scene occlusion, and container-based concealment and movement; the bottom row evaluates object solidity through obstruction, support removal, and collision. Each panel shows representative frames from a sample drawn from one generator in the corresponding task family: an early input frame, the final input frame at the split boundary, and a target frame depicting the expected physical outcome.}\label{fig:task-taxonomy}\end{figure*}

\vspace{-3mm}
 
\subsection{Cognitive Taxonomy}
\label{sec:taxonomy}
 
\begin{wrapfigure}{r}{0.49\textwidth}\vspace{-12pt}\centering\includegraphics[width=0.6\linewidth]{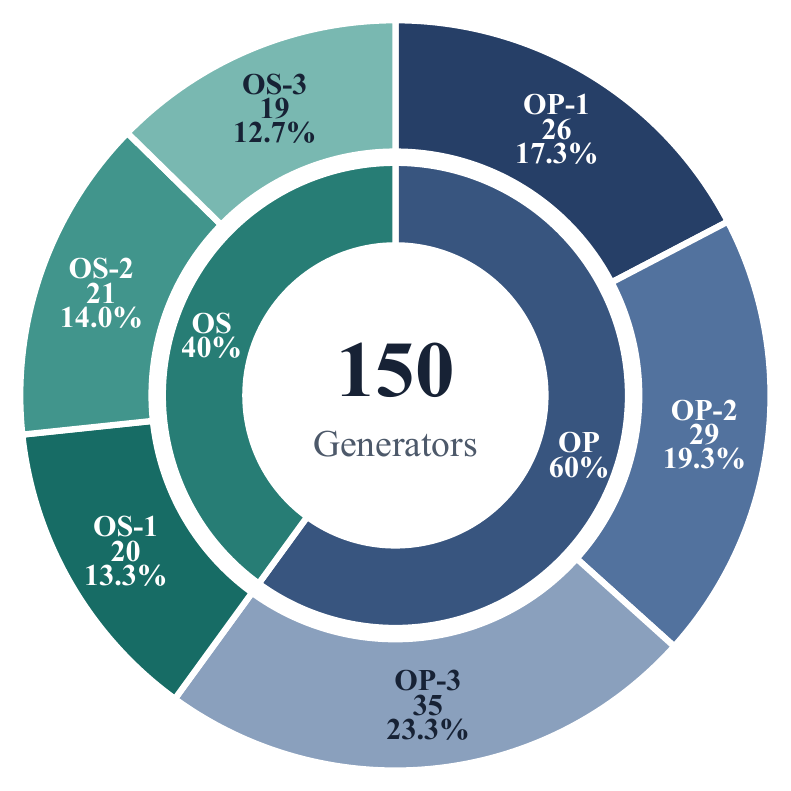}\caption{Generator-level composition of WROP's 150 generators. The inner ring shows 90 OP and 60 OS generators; the outer ring shows the six task families.}\label{fig:taxonomy-composition}\vspace{-8pt}\end{wrapfigure}

WROP organizes \textbf{150 task generators} into six families across two cognitive dimensions \citep{spelke2007core,baillargeon1986representing,hespos2001reasoning,sanford1967volume}. OP tasks require the model to maintain and reinstate object representations across periods of occlusion; OS tasks require it to generate the mechanical consequences of solid boundaries. The six families are illustrated in Figure~\ref{fig:task-taxonomy}, and their generator-level composition is summarized in Figure~\ref{fig:taxonomy-composition}.

Each task family is designed to probe a specific aspect of object permanence or object solidity, adapted from established experimental paradigms where applicable and otherwise constructed originally to suit the demands of video generation evaluation. Every generator is authored so that the key physical event begins at or after the \textbf{temporal midpoint}: the \textit{input} half establishes pre-event scene context and the \textit{target} half captures the event and its physical consequence, aligning with the V2V evaluation protocol. Within each generator, parameters are partitioned into two sets, both varied to promote sample diversity but serving distinct roles. \textbf{Structural parameters}, including but not limited to object count, geometry, trajectory, occlusion configuration, aperture size, and contact timing, define the physical and cognitive challenge; they are varied systematically across samples within a generator to control task difficulty and ensure that models cannot succeed by memorizing a fixed physical outcome. \textbf{Surface-level parameters}, including but not limited to object color, material, scene lighting, and camera viewpoint, are randomized independently across samples within the same structural configuration to maximize visual diversity without altering the underlying physical problem, preventing models from exploiting perceptual cues in place of physical reasoning.
 
\begin{description}
\setlength{\itemsep}{6pt}

\item[\textbf{OP-1:} \texttt{Baillargeonian\_Occlusion}.\footnotemark]
A target object moves along a defined trajectory and passes behind an occluder; the model must generate its re-emergence on the distal side with identity, size, and motion direction intact \citep{baillargeon1986representing}. This probes whether the model maintains a persistent object representation through complete visual absence rather than extrapolating motion from the last visible frame. \textbf{Structural parameters:} object count, track topology (linear, curved, multi-pass), occluder opacity, and occlusion duration.

\item[\textbf{OP-2:} \texttt{Object\_Static\_Occlusion}.]
A moving occluder covers a known static configuration of objects; upon removal, the scene must be reinstated with number, identity, and spatial arrangement unchanged \citep{stahl2015observing,wynn1992addition}. This probes the representation of multiple hidden objects simultaneously: the model must treat occlusion as causally inert rather than as an event that transforms the hidden scene. \textbf{Structural parameters:} occluder motion type (translational, rotational, split-panel), coverage fraction, object count, number of panels, and reveal dynamics.

\item[\textbf{OP-3:} \texttt{Container\_Permanence}.]
An object is concealed inside a container that may remain static or undergo displacement, rotation, or swapping among alternatives; the model must generate the object as bound to its container's new position rather than its world-origin location \citep{hespos2001reasoning}. This probes spatial reference-frame updating under containment: a more demanding form of permanence in which location must be continuously recomputed as a function of a moving reference object. \textbf{Structural parameters:} container state (static or dynamic), closure mechanism, number of containers and swap events, object count, and path complexity.

\item[\textbf{OS-1:} \texttt{Baillargeonian\_Obstruction}.$^{1}$]
A moving object approaches a barrier whose aperture is either smaller than the object (blocking) or larger (permitting); the model must generate the physically correct outcome for each case \citep{baillargeon1985object,hood2000predicting}. This probes geometric solidity reasoning: the model must evaluate the spatial relationship between object size and aperture size to determine whether passage is physically possible, rather than defaulting to a prior that objects in motion continue moving. \textbf{Structural parameters:} barrier type (planar, angled, compound, multi-layer), object-to-aperture size ratio, approach speed, and barrier visibility.

\item[\textbf{OS-2:} \texttt{Object\_Drop}.]
A support surface is withdrawn from beneath an object, which must then fall; a size-aperture filter below determines whether the object passes through a lower surface or comes to rest upon it \citep{hespos2012physics}. This probes support-contingent gravity: the model must couple the onset of falling to the removal of support rather than applying continuous downward motion or leaving the object suspended. \textbf{Structural parameters:} drop mechanism (instantaneous removal, gradual withdrawal, causal chain), object type, causal chain visibility, and size-filter configuration.

\item[\textbf{OS-3:} \texttt{Object\_Collision}.]
A moving object strikes a stationary configuration; the model must generate physically consistent post-collision trajectories for all objects while preserving count and identity throughout \citep{sanford1967volume}. This probes contact-mediated solidity: objects must neither merge, annihilate, nor pass through one another on impact, and momentum transfer must produce diverging rather than coincident trajectories. \textbf{Structural parameters:} number of objects, collision geometry (direct, glancing, chain transfer), number of stationary intermediaries, impact symmetry, and post-collision trajectory complexity.

\end{description}

\footnotetext{OP-1 and OS-1 are directly modelled on Ren\'{e}e Baillargeon and colleagues' seminal experimental schematics \citep{baillargeon1986representing,baillargeon1985object}; their names pay homage to this foundational lineage. However, unlike the original tasks, which employed the violation-of-expectation (VoE) paradigm---in which an impossible event is presented to elicit gaze orienting---to probe physical reasoning in pre-verbal infants, video generation models are here asked to produce the physically plausible continuation directly.}
 
\subsection{Data Statistics}
\label{sec:stats}
 
The training corpus contains {1,500,000 samples} across {150} generators, each contributing {10,000} samples. The evaluation exam contains {300 questions}: {2} samples from each of the 150 generators. Every sample is a 120-frame, physically consistent, hand-authored animation rendered at 1280$\times$720 and 24\,fps, split at the onset of the key event into a {60-frame} \textit{input video} and a {60-frame} \textit{target video}, together with a natural-language \textit{prompt}, a per-frame \textit{trajectory} of object poses, and a \textit{metadata} record describing the scene state. Motion is authored as Blender keyframe animation rather than produced by a physics engine; the trajectory arrays are sampled from that animation.
 
\subsection{Data Generation Pipeline}
\label{sec:pipeline}
 
\begin{figure}[h] \centering \includegraphics[width=\textwidth]{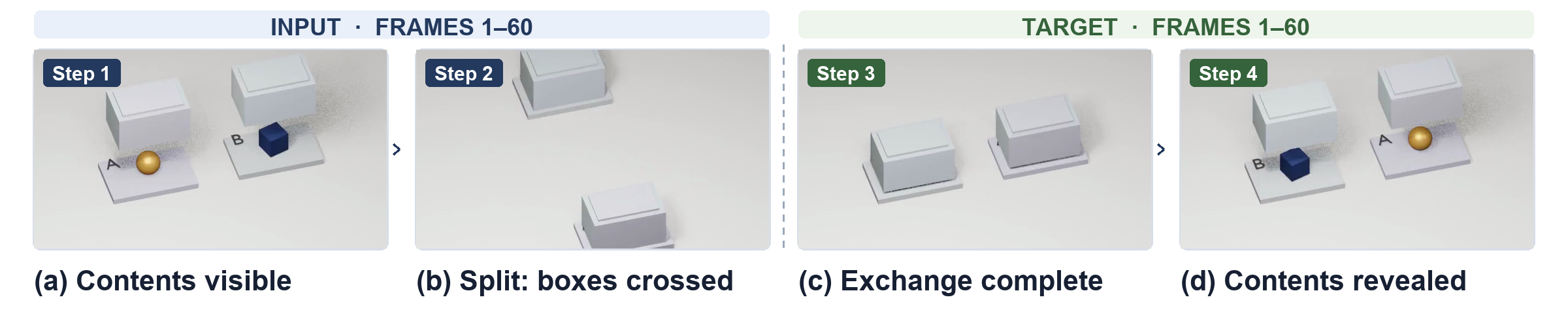} \caption{Generator-design example from \texttt{Marked\_Boxes\_Swap} using the current 60-frame input and 60-frame target clips. The sequence shows the objects in marked boxes, the closed boxes after crossing at the split, completion of the position exchange, and the identity-preserving reveal. The task requires the model to maintain the association between each hidden object and its marked container across the input--target boundary rather than infer identity from final screen position.} \label{fig:marked-boxes-swap-design} \end{figure}

\paragraph{Generator design.} Each generator instantiates its task family's physical scenario as a self-contained 3D Blender scene. Diverse everyday objects and scene configurations are used to test the same physical principle across visually distinct settings. Where permitted by the task, we introduce multiple physically valid outcomes within a single generator: in \texttt{Marked\_Boxes\_Swap}, for example, two labeled boxes close over distinct objects, exchange screen positions, and reopen with each object still associated with its original marked box. This construction prevents models from using final position alone and instead requires them to track box identity and hidden contents through motion and occlusion. Scene geometry, object trajectories, contact timing, occlusion coverage, and camera placement are revised whenever object interpenetration or other physical violations are observed during inspection. Figure~\ref{fig:marked-boxes-swap-design} illustrates this design principle. The data generation pipeline consists of three stages.

\textbf{(1) Task-specific generator implementation.}
Each of the 150 tasks is implemented as a self-contained, parameterized Blender generator specifying objects and their semantic roles, scene geometry, initial conditions, the keyframed motion and contact events that constitute the task, camera configuration, natural-language prompt, and expected physical outcome. No rigid-body solver is used: every trajectory is authored analytically so that occlusion, contact, and reappearance occur at controlled frames, and physical plausibility is the responsibility of the scene author rather than of a simulator.
 
\textbf{(2) Sample generation and construction.}
A shared driver executes each generator through a common Blender rendering backend (Blender 4.4.3, EEVEE Next). A recorded random seed controls surface variations in object color, material, and scene lighting, while the authored camera, geometry, spatial configuration, and physical mechanism are preserved. The renderer produces a 120-frame animation, split at frame 60 to yield a 60-frame \textit{input video} and a 60-frame \textit{target video}. Each sample is packaged as a five-tuple: \textit{input video}, \textit{target video}, \textit{prompt}, \textit{trajectory}, and \textit{metadata}.
 
\textbf{(3) Large-scale generation and validation.}
Generators run independently across parallel workers. Each contributes 10,000 training samples; failed renders are automatically retried and logged, and an automated audit verifies file completeness and schema validity. Each generated sample undergoes automated validation before admission to the dataset. We verify that all five components (\textit{input video}, \textit{target video}, \textit{prompt}, \textit{trajectory}, \textit{metadata}) are present and readable, that both clips share the same frame rate, and that each contains exactly {60 frames} meeting at {frame 60} without a gap or overlap. Trajectory and metadata files are checked for required fields; the metadata records generator identity, sample index, random seed, and rendering configuration, allowing every sample to be traced to its generation conditions. Samples failing any check are rejected and regenerated. Before release, representative samples from every generator are {manually inspected} to confirm that the rendered sequence matches the intended task definition and that the split boundary is correctly placed. 

\section{Evaluation}
\label{sec:evaluation}

Here we ask two questions: how well current video models respect object permanence and solidity, and whether these principles can be trained in with a core-cognition dataset. For the second question we fine-tune PWM-WROP, a 16B open-weight world model, on the WROP training corpus, so that it serves as a baseline for the corpus rather than a new model design (Section~\ref{sec:pwm-wrop}; its training stack, including a native-PyTorch implementation for AWS Trainium2, is documented in Appendix~\ref{app:infra}). For the first question we evaluate PWM-WROP alongside thirteen open-weight and proprietary systems that fall into three interface classes: true continuation, reference-to-video, and edit or transfer (Section~\ref{sec:models}). All fourteen are driven by one inference harness that passes each question's input video and prompt verbatim and never exposes the target (Section~\ref{sec:protocol}). Our primary measure is human preference from blind pairwise comparisons fitted with a Bradley--Terry model (Section~\ref{sec:human}); a suite of full-reference metrics against the target video serves as a secondary, target-fit measure (Section~\ref{sec:auto-metrics}). Results follow in Section~\ref{sec:results}.

\subsection{PWM-WROP}
\label{sec:pwm-wrop}

\textbf{Base model and objective.} PWM-WROP is fine-tuned from Cosmos3-Nano \citep{nvidia2026cosmos3}, with only the training signal changed; the architecture and tokenizer remain identical to the base model. The model is trained on the WROP video-to-video contract itself: the input half of a sample is the conditioning clip and the target half is the prediction target, so the key physical event and its consequence always fall in the frames the model must generate (Section~\ref{sec:pipeline}). Training and evaluation thus operate on a shared protocol, and no task labels, family names, or supervision beyond the clip and its natural-language prompt are used. The model is fine-tuned for one epoch on 1,500,000 samples, an earlier render of the same 150 generators, with each sample's prompt as text conditioning. The training geometry is a 117-frame packed clip at $320\times192$ (57 conditioning frames followed by 60 predicted frames), which is also the geometry the model is evaluated at (Section~\ref{sec:protocol}). PWM-WROP is the only model in this study trained on WROP data.

\textbf{Training stack.} PWM, the stack that produced the checkpoint, is released alongside it. Its native-PyTorch implementation for AWS Trainium2 shards the 36-layer model over a tensor-parallel $\times$ FSDP2 mesh of 64 NeuronCores, compiles each block once with static shapes, and reaches 5.7\,s per step of batch 16 at the $288\times512$ geometry. Appendix~\ref{app:infra} gives the parallel layout, the throughput ledger, the engineering findings that carried the step from 15.1\,s to 5.7\,s, the correctness gates the port passed, and the status of the end-to-end fine-tune of PWM-WROP on said infrastructure.

\subsection{Models Evaluated}
\label{sec:models}

We evaluate fourteen video-to-video models on WROP: PWM-WROP (our model) and thirteen external systems spanning open-weight and proprietary families (Table~\ref{tab:models}). Open-weight models run locally on our hardware; proprietary models are accessed through hosted APIs. Every model receives the same two inputs per benchmark item---the conditioning \textit{input video} and the natural-language \textit{prompt}---and is tasked with producing the continuation. The \textit{target video} is withheld from all models throughout evaluation.

The fourteen systems fall into three interface classes that reflect fundamentally different relationships to the conditioning clip. \emph{True continuation} models including PWM-WROP, MAGI-1 24B, LTX-2.3 Extend, and Grok Imagine (via its video-extend endpoint) treat the source clip as a prefix and synthesize the frames that follow it. \emph{Reference-to-video} models including Seedance 2.5, Wan 3.0 Prime, and MiniMax H3 treat the source as a visual reference and regenerate the full event from the prompt on their own timeline, producing a fresh rendition rather than a temporal extension. \emph{Edit and transfer} models including Wan-VACE 14B, HY-OmniWeaving, LTX-2.3 Dev with IC-LoRA conditioning, Cosmos3 Super, Kling O3 Pro, Gemini Omni Flash 1.1, and Runway Aleph 2 repaint the source span frame by frame, so their output occupies the same temporal interval as the input and cannot depict events that unfold after the occlusion boundary. Because these three classes engage the benchmark under qualitatively different assumptions, we treat interface class as an explicit factor in all subsequent analyses.

\begin{table}[t]
\centering
\small
\setlength{\tabcolsep}{5pt}
\resizebox{\textwidth}{!}{\begin{tabular}{llll}
\toprule
\textbf{Model} & \textbf{Access} & \textbf{Interface class} & \textbf{Native output} (res / fps / frames) \\
\midrule
\multicolumn{4}{l}{\textit{Ours}} \\
PWM-WROP & Trainium2 48XL & True continuation & $320\times192$ / 24 / 60 predicted$^{\dagger}$ \\
\midrule
\multicolumn{4}{l}{\textit{Open-weight}} \\
MAGI-1 24B & Local, multi-GPU & True continuation & $1280\times720$ / 24 / 77 \\
LTX-2.3 Dev & Local, multi-GPU & Edit / transfer (IC-LoRA) & $1536\times1024$ / 24 / 121 \\
Wan-VACE 14B & Local, multi-GPU & Edit / transfer (repaint) & $832\times480$ / 16 / 57 \\
HY-OmniWeaving & Local, multi-GPU & Edit / transfer (editing) & $848\times480$ / 24 / 57 \\
Cosmos3 Super & Local, multi-GPU & Edit / transfer (edge control) & $1280\times720$ / 24 / 60 \\
\midrule
\multicolumn{4}{l}{\textit{Proprietary}} \\
LTX-2.3 Extend & FAL API & True continuation & $1920\times1080$ / 24 / 145$^{\ddagger}$ \\
Grok Imagine (video extend) & FAL API & True continuation & $1280\times720$ / 24 / 132$^{\ddagger}$ \\
Seedance 2.5 & FAL API & Reference-to-video & $1280\times720$ / 24 / 97 \\
Wan 3.0 Prime & FAL API & Reference-to-video & $1280\times720$ / 30 / 90 \\
MiniMax H3 & FAL API & Reference-to-video & $1344\times768$ / 24 / 124 \\
Kling O3 Pro & FAL API & Edit / transfer (frame-wise) & $1920\times1080$ / 24 / 73$^{\ddagger}$ \\
Gemini Omni Flash 1.1 & FAL API & Edit / transfer (frame-wise) & $1280\times720$ / 24 / 60 \\
Runway Aleph 2 & Runway API & Edit / transfer (frame-wise) & $1920\times1080$ / 24 / 60 \\
\bottomrule
\end{tabular}}
\caption{The fourteen evaluated models, grouped by provenance, with access route, interface class and native output geometry. $^{\dagger}$Our model emits a 117-frame clip that replays its 57 conditioning frames before the 60 predicted frames; the replay is trimmed before evaluation. $^{\ddagger}$Stitched or padded outputs whose source prefix is trimmed before evaluation (Section~\ref{sec:protocol}).}
\label{tab:models}
\end{table}

\subsection{Inference Protocol}
\label{sec:protocol}

All fourteen models are evaluated through a unified inference harness. Each benchmark item comprises a \textit{prompt}, a conditioning \textit{input video}, and a held-back \textit{target video}; the harness passes only the first two to the model, writes the output to a standardized location, and admits it only after \texttt{ffprobe} confirms a decodable video with the expected geometry. No model-specific prompt engineering is applied: prompts are transmitted verbatim without chain-of-thought framing or task labels, and server-side prompt expansion is disabled on every endpoint that exposes the option.

\textbf{Open-weight models} run in isolated environments on local hardware, each with its own dependency set and checkpoint, at the resolution and sampler settings recommended upstream. Smaller editors run on a single GPU; MAGI-1 24B and Cosmos3 Super require multi-GPU nodes. PWM-WROP conditions on the last 57 frames of the input and predicts 60 frames at $320\times192$ with a fixed seed (UniPC, 35 steps, guidance 6.0, shift 10.0). For the two 90-frame benchmark items, the first 33 input frames fall outside the model's conditioning window and are not seen.

\textbf{Proprietary models} are called through hosted HTTP APIs, all but Runway Aleph 2 through a single aggregation provider. The requested extension length is derived per item from the target's frame count, so that 90-frame items receive a 3.75\,s request rather than 2.5\,s. Inputs shorter than a provider's minimum clip length are front-padded by repeating the first frame, which preserves the timing of the event on which permanence and solidity are judged; the padded duration is recorded per generation. Endpoints that return the source stitched to their output are trimmed to the predicted span before evaluation: LTX-2.3 Extend drops its first 3.25\,s (0.75\,s of padding plus the 2.5\,s source; 3.75\,s for 90-frame items), Grok Imagine drops its first 60 frames (90 for 90-frame items), and Kling O3 Pro drops its 0.5\,s front pad. The full request payload, returned geometry, and any padding are recorded per generation.

\subsection{Human Judgment}
\label{sec:human}

Human preference is our primary evaluation measure, collected as pairwise blind comparisons across all fourteen models.

\textbf{Setup.} All clips were normalized to $1280\times720$, 24\,fps, silent, at a common bitrate before rating. Models whose outputs replay or pad the source were trimmed to their predicted span (Section~\ref{sec:protocol}). Raters were shown the \textit{input video} and the text prompt, then presented with two anonymized, randomly ordered continuations (A and B) and asked to select A, B, or ``about the same.'' Judgments were made on three criteria jointly: alignment with the text description, natural motion and physical plausibility, and object permanence---specifically, that objects do not vanish, appear spontaneously, pass through solid barriers, or change in color, shape, or count following occlusion.

\textbf{Quality control.} Twenty crowdsourced raters participated after passing a qualification screen (threshold: 8/10; median score: 9/10). Raters answered 90.0\% of embedded attention checks correctly and agreed with themselves on 93.5\% of repeated items (Cohen's $\kappa = 0.891$). No position bias was detected: the left-placed clip was preferred in 51.4\% of non-tie judgments ($p = 0.615$).

\textbf{Scoring.} Each of the 120 candidate model pairs was scheduled across 4 benchmark items, yielding 476 of 480 completed judgments, of which 361 fell between distinct models (50--52 per model). Pairwise outcomes are aggregated via a Bradley--Terry model with ties scored as half-wins; we report Elo-scale strengths with a global mean of 1,500. Confidence intervals and top-1 probabilities are derived from 1,000 bootstrap replicates resampling over raters. Repeated items are excluded from the primary analysis; including them leaves the top-ten ranking unchanged.

\subsection{Automatic Metrics}
\label{sec:auto-metrics}

As a secondary complement to human judgment, we compute a suite of full-reference metrics against the \textit{target video}: pixel-level (MSE, MAE), signal fidelity (PSNR, SSIM, MS-SSIM), perceptual (LPIPS), temporal (temporal-difference L1), and final-frame variants of MSE, SSIM, and LPIPS alongside CLIP similarity and FID. Each model's predicted span is resampled to the target's frame count and resized to $320\times192$, the lowest native resolution in the evaluation pool, to equate spatial scale across models. Final-frame metrics are additionally computed at $1280\times720$, though resolution differences across providers remain a confound at that scale. These metrics quantify proximity to the reference clip rather than physical reasoning correctness. An edit model that faithfully repaints the static pre-event scene, for instance, may score well on SSIM without ever depicting the object re-emerging from occlusion. For this reason, automatic metrics are treated as secondary throughout, and the computation scale is stated alongside every reported value.

\section{Results}
\label{sec:results}

\subsection{Human Preference}
\label{sec:res-human}

\subsubsection{Overall Results}

\begin{figure*}[h]
\centering
\includegraphics[width=0.9\textwidth]{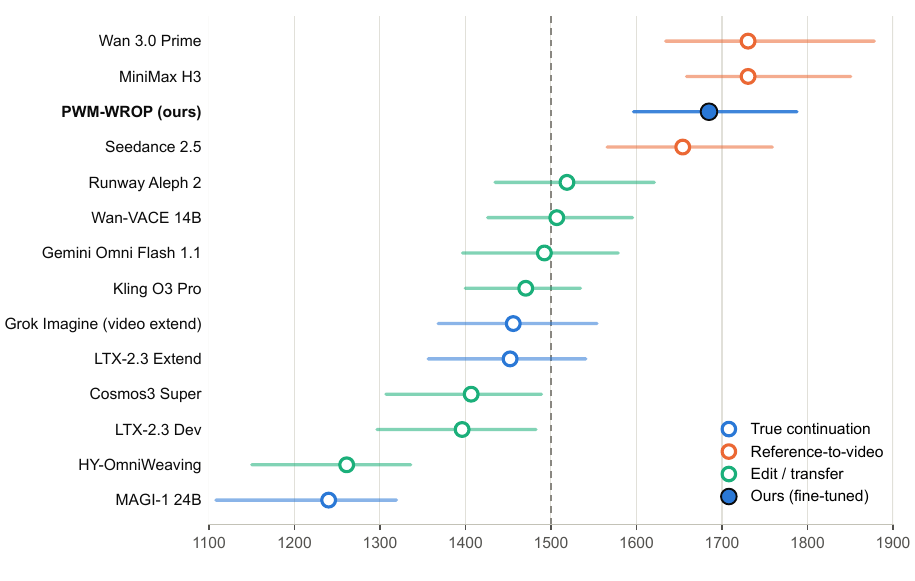}
\caption{Human-preference Elo of the 14 models. Points are Bradley--Terry strengths on an Elo scale (mean 1500, dashed line), bars are 95\% rater-clustered bootstrap intervals, colour is interface class, and PWM-WROP (filled) is the top true-continuation model.}
\label{fig:elo}
\end{figure*}

Table~\ref{tab:elo} and Figure~\ref{fig:elo} report Bradley--Terry strengths on an Elo scale (mean 1500), fitted on 361 pairwise judgments across all fourteen models (50--52 judgements per model), with 95\% rater-clustered bootstrap intervals from 1{,}000 replicates (Section~\ref{sec:human}). The fourteen systems span three structurally distinct interface classes (true continuation, reference-to-video, and edit/transfer) that engage the benchmark under qualitatively different assumptions (Section~\ref{sec:models}).

\begin{table}[h]
\centering
\caption{Human preference leaderboard of the 14 evaluated models on the 300-question WROP exam. Twenty raters made 361 blind pairwise judgments between the fourteen models (50--52 per model; ties count 0.5 for each side); strengths are Bradley--Terry maximum-likelihood estimates, with one virtual draw added per model and distributed evenly across its opponents, rescaled to an Elo scale with mean 1500. 95\% confidence intervals come from 1{,}000 rater-clustered bootstrap resamples; overlapping intervals should not be read as significant rank differences. Score rate is the raw win rate (win~$=1$, tie~$=0.5$). Wan 3.0 Prime and MiniMax H3 have identical records and tie for first.}
\label{tab:elo}
\small
\setlength{\tabcolsep}{4pt}
\begin{tabular}{rllrcrr}
\toprule
Rank & Model & Class & Elo & 95\% CI & Score rate & Games \\
\midrule
1 & Wan 3.0 Prime & Reference-to-video & 1723.6 & [1629.2, 1864.9] & 77.9\% & 52 \\
2 & MiniMax H3 & Reference-to-video & 1723.6 & [1644.5, 1837.6] & 77.9\% & 52 \\
\textbf{3} & \textbf{PWM-WROP (ours)} & \textbf{True continuation} & \textbf{1679.5} & \textbf{[1603.5, 1781.5]} & \textbf{73.1\%} & \textbf{52} \\
4 & Seedance 2.5 & Reference-to-video & 1649.6 & [1554.2, 1751.5] & 69.6\% & 51 \\
5 & Runway Aleph 2 & Edit / transfer & 1518.3 & [1425.1, 1621.6] & 52.9\% & 51 \\
6 & Wan-VACE 14B & Edit / transfer & 1506.7 & [1429.9, 1590.8] & 51.0\% & 52 \\
7 & Gemini Omni Flash 1.1 & Edit / transfer & 1492.5 & [1404.1, 1572.9] & 49.0\% & 52 \\
8 & Kling O3 Pro & Edit / transfer & 1471.3 & [1404.0, 1534.0] & 46.2\% & 52 \\
9 & Grok Imagine (video extend) & True continuation & 1457.0 & [1362.7, 1555.1] & 44.2\% & 52 \\
10 & LTX-2.3 Extend & True continuation & 1453.4 & [1363.6, 1545.8] & 43.1\% & 51 \\
11 & Cosmos3 Super & Edit / transfer & 1409.2 & [1318.2, 1488.7] & 37.3\% & 51 \\
12 & LTX-2.3 Dev & Edit / transfer & 1398.9 & [1297.7, 1482.3] & 36.5\% & 52 \\
13 & HY-OmniWeaving & Edit / transfer & 1268.5 & [1159.4, 1344.2] & 21.0\% & 50 \\
14 & MAGI-1 24B & True continuation & 1248.0 & [1137.2, 1315.6] & 19.2\% & 52 \\
\bottomrule
\end{tabular}
\end{table}

\textbf{Reference-to-video models dominate the overall leaderboard.} Wan~3.0 Prime and MiniMax~H3 share the top two positions at 1724 each, and Seedance~2.5 ranks fourth at 1650. Unlike true-continuation or edit/transfer models, reference-to-video systems do not have to continue directly from the end of the input clip. Instead, they regenerate the scene and its continuation on their own timeline. This difference may help explain their strong performance. Because they generate the continuation from scratch rather than extending the input from its final frame, they have more freedom to produce a physically coherent scene. However, this freedom can also cause problems: when the model needs to preserve the input video's object identities and spatial arrangements, regenerating the scene may cause it to follow its own interpretation rather than preserve what was originally shown. (see Seedance~2.5's failures on G19 (OP-2) and G27 (OS-2) in Section~\ref{sec:qualitative} for details).

\textbf{Fine-tuning on domain-specific data leads to a substantial improvement among true-continuation models.} PWM-WROP ranks third overall at 1680 [1604, 1782] and is the highest-ranked true-continuation model, leading the next-best true-continuation system, Grok Imagine (video extend), by 224 Elo points (1457 [1368, 1555]). The remaining true-continuation models, LTX-2.3 Extend (1452) and MAGI-1 24B (1240), rank lower still. This result supports the idea that fine-tuning with concept-specific synthetic data can improve performance on object permanence and solidity reasoning. It also suggests that targeted training could be useful for improving V2V models on these capabilities. However, because the models also differ in architecture, the performance gap cannot be attributed to training alone.

\textbf{Interface type explains the leaderboard better than model scale.} Bootstrap top-1 probabilities show a clear gap between the top four systems and the rest: Wan~3.0 Prime leads at 46.8\%, followed by MiniMax~H3 at 36.4\%, PWM-WROP at 12.3\%, and Seedance~2.5 at 4.5\%; every other system has zero probability of ranking first. The five systems from Runway Aleph~2 through Cosmos3 Super are separated by only 109 Elo points, with overlapping intervals that make it difficult to distinguish them statistically. In contrast, HY-OmniWeaving and MAGI-1 24B fall well below this group. The middle of the leaderboard is made up almost entirely of edit/transfer models, with two true-continuation systems mixed in. One possible explanation is that the frame-level repainting used by edit/transfer models makes it harder for them to generate events that happen after the input ends, causing these models to perform similarly and cluster together on the leaderboard, regardless of differences between individual models. Proprietary models (e.g. Gemini Omni Flash~1.1 and Kling O3 Pro) also appear in the middle of the leaderboard alongside open-weight models, while Wan-VACE 14B performs similarly to other models in the middle of the leaderboard despite its relatively small disclosed parameter count.

\subsubsection{Performance by Task Family}
\label{sec:res-family}

\begin{figure*}[h]
\centering
\includegraphics[width=\textwidth]{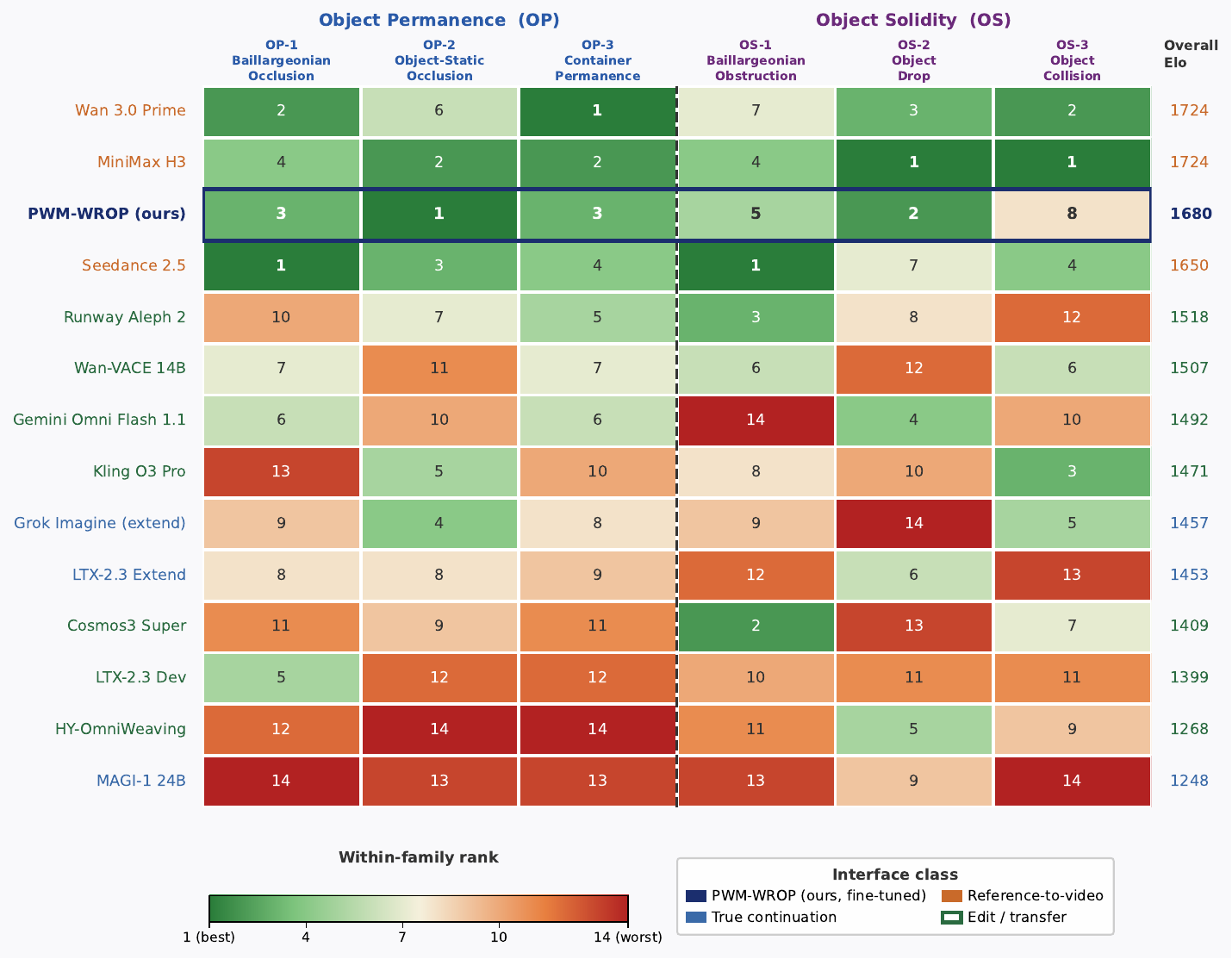}
\caption{Within-family human-preference ranks across the six WROP task families. Each cell shows the Bradley--Terry rank within that family (1 = best, 14 = worst), fitted separately on the 20-rater pairwise judgments for each family (36--96 games per family; 361 total). Color encodes rank from green (top) to red (bottom). The outlined row marks PWM-WROP; overall Elo (right) is from the joint fit over all judgments. Row order follows overall Elo rank; column groups correspond to the task families in Figure~\ref{fig:task-taxonomy}. Interface class is indicated by the label color, matching Figure~\ref{fig:elo}.}
\label{fig:family-heatmap}
\end{figure*}

We further break down performance by task family to examine model capabilities across different cognitive concepts. Figure~\ref{fig:family-heatmap} shows within-family Bradley--Terry ranks across the six task families for all fourteen models. The analysis reveals a clear difference across PWM-WROP’s task performance: it ranks first in object-static occlusion (OP-2) and third in both Baillargeonian occlusion (OP-1) and container permanence (OP-3); across the three OS families its profile is more variable, placing second in object drop (OS-2), fifth in Baillargeonian obstruction (OS-1), and eighth in object collision (OS-3). This difference between OP and OS performance suggests that the two capacities may require different internal representations \citep{falck2020core}, and that the current fine-tuning regime is more effective for occlusion tracking than for contact-based dynamics.

At the same time, the two reference-to-video leaders derive much of their overall advantage from OS families: MiniMax~H3 ranks first in both object drop and collision, winning 100\% of its games in each; Wan~3.0 Prime likewise leads in container permanence and ranks second in collision. Their OP rankings are comparatively moderate, suggesting that the greater freedom of reference-to-video models to regenerate a scene may be particularly helpful for solidity tasks but less helpful for occlusion tracking, where the model must more strictly preserve the objects and spatial arrangements established in the input.

\subsection{Qualitative Analysis}
\label{sec:qualitative}

We examine model behavior at the task level through same-task, same-sample comparisons on six representative generators, one from each family. The cases are drawn from generators that have major effects on the family-level Elo results (Section~\ref{sec:res-family}). This qualitative analysis helps explain the quantitative results and helps identify where and how models fail.

\begin{figure*}[h]
\centering
\includegraphics[width=\textwidth]{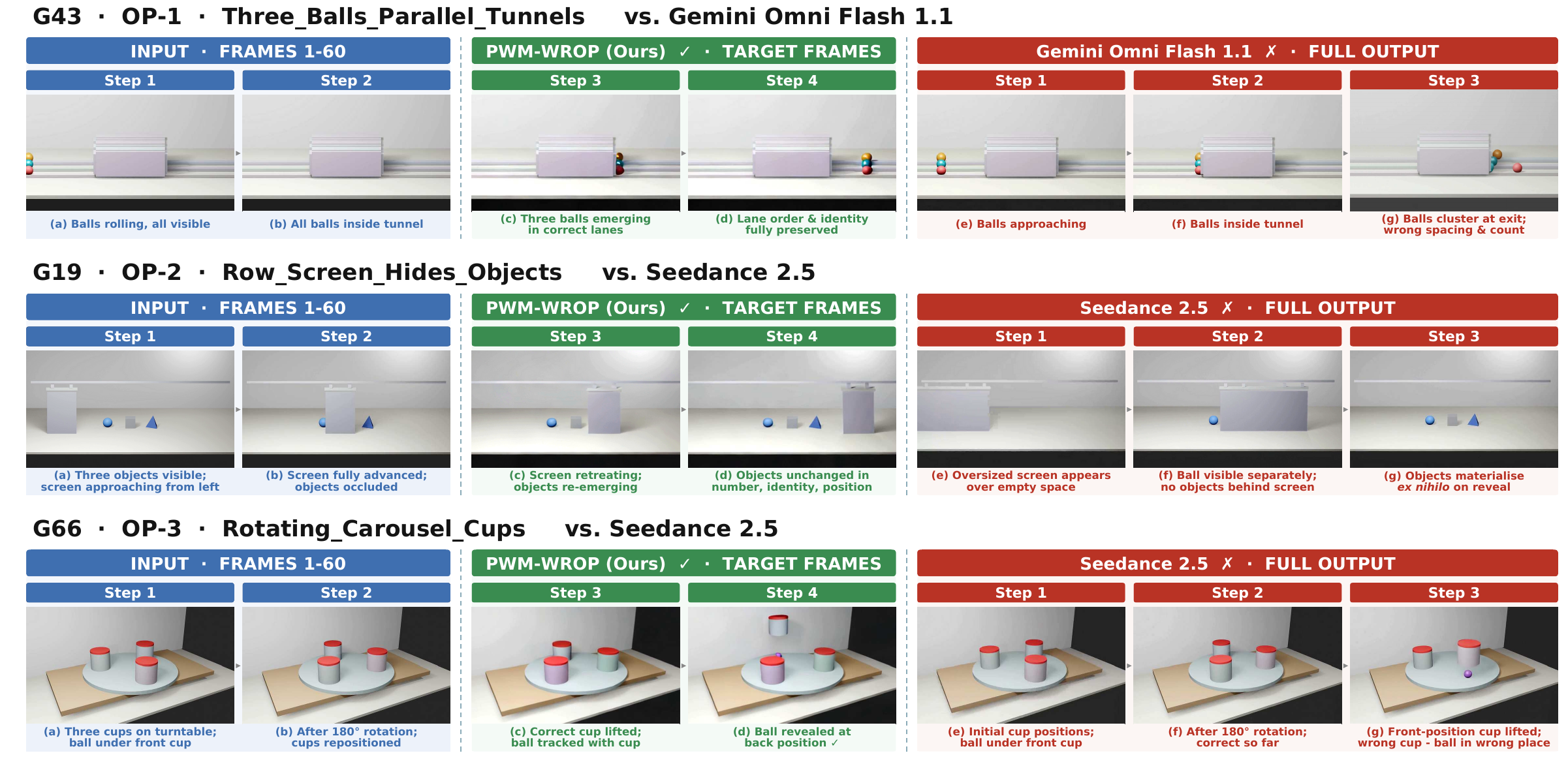}
\caption{Qualitative comparison on three object-permanence task families (Figure~\ref{fig:task-taxonomy}, top row). Each row shows the same ground-truth input (two key frames, blue), the PWM-WROP target output (two frames, green), and the full baseline output (three frames spanning the generation, red), with \textcolor{green}{$\checkmark$}/\textcolor{red}{$\times$} badges indicating physical correctness. \textbf{G43 (OP-1):} Gemini Omni Flash collapses all three balls into a tight cluster at the tunnel exit, violating their lane identities and count, while PWM-WROP keeps each ball in its correct lane with continuous motion. \textbf{G19 (OP-2):} Seedance 2.5 generates an oversized occluder over an empty region of the scene while the original objects remain visible, then makes all three objects suddenly appear when the occluder moves away, with no clear continuity from their previous positions; PWM-WROP correctly preserves the scene and reveals the same unchanged configuration. \textbf{G66 (OP-3):} Seedance 2.5 correctly animates the 180\textdegree{} turntable rotation but then lifts the cup at the original front position--a different physical cup--rather than tracking the cup containing the ball to its new location; PWM-WROP correctly tracks the ball with the rotating cup and lifts the correct cup.}
\label{fig:qualitative-op}
\end{figure*}

\textbf{Identity and count preservation through occlusion (G43, OP-1).} \texttt{Three\_Balls\_Parallel\_Tunnels} shows three colored balls rolling in parallel lanes into an opaque tunnel. Given the input video, the model must generate a target video that continues, in which all three balls come out on the other side in the same lane order, with the same colors and total count. PWM-WROP correctly keeps track of each ball while it is hidden and brings all three back in their original lanes. Gemini Omni Flash, which edits the source clip frame by frame, fails to preserve the three balls during occlusion: its final output contains an extra fourth ball, and the balls are no longer in the correct lanes or spacing. This is a clear object-permanence failure: the model does not maintain “what is where” during occlusion and instead generates an exit event that is inconsistent with the balls' trajectories before the occlusion.

\textbf{Persistence of a hidden scene under moving occlusion (G19, OP-2).} \texttt{Row\_Screen\_Hides\_Objects} shows three distinct objects in a row while a screen moves in front of them. The model must continue the video until the screen moves away and the three same objects are visible again in the same arrangement. PWM-WROP correctly treats the screen as blocking the objects from view and reveals the unchanged arrangement when the screen moves away. In contrast, Seedance 2.5 exposes two related problems that violate the object-permanence principle. First, it generates an oversized screen that moves through a region containing no objects. The original objects remain partially visible next to the screen rather than behind it, suggesting that the model does not correctly align the screen’s position and size with the objects it should conceal. Second, when the screen moves away, all three objects suddenly appear from an empty area, with no clear continuity from their pre-occlusion positions. This suggests that the model does not keep the objects present while they are hidden, but instead makes them disappear and then generates them again when the screen moves away.

\textbf{Reference-frame updating under container displacement (G66, OP-3).} \texttt{Rotating\_Carousel\_Cups} places three identical cups on a rotating turntable with a ball concealed under the front cup before rotation; after a 180\textdegree{} turn, the model must lift the cup that began at the front, now positioned at the back, to reveal the ball. This task requires the model to remember which cup the ball is under and update the ball’s location when that cup moves. PWM-WROP correctly updates the ball's world position as the turntable rotates, lifting the cup at the back and revealing the ball at the correct position under the original cup. Seedance 2.5 correctly generates the turntable rotation but then opens the cup at the front instead of the cup that originally covered the ball, revealing the ball under the front cup instead. This mistake suggests that Seedance 2.5 tracks the ball by its position on the turntable rather than by the cup covering it, so when that cup moves to the back, the model fails to move the ball with it.

\begin{figure*}[h]
\centering

\includegraphics[width=\textwidth]{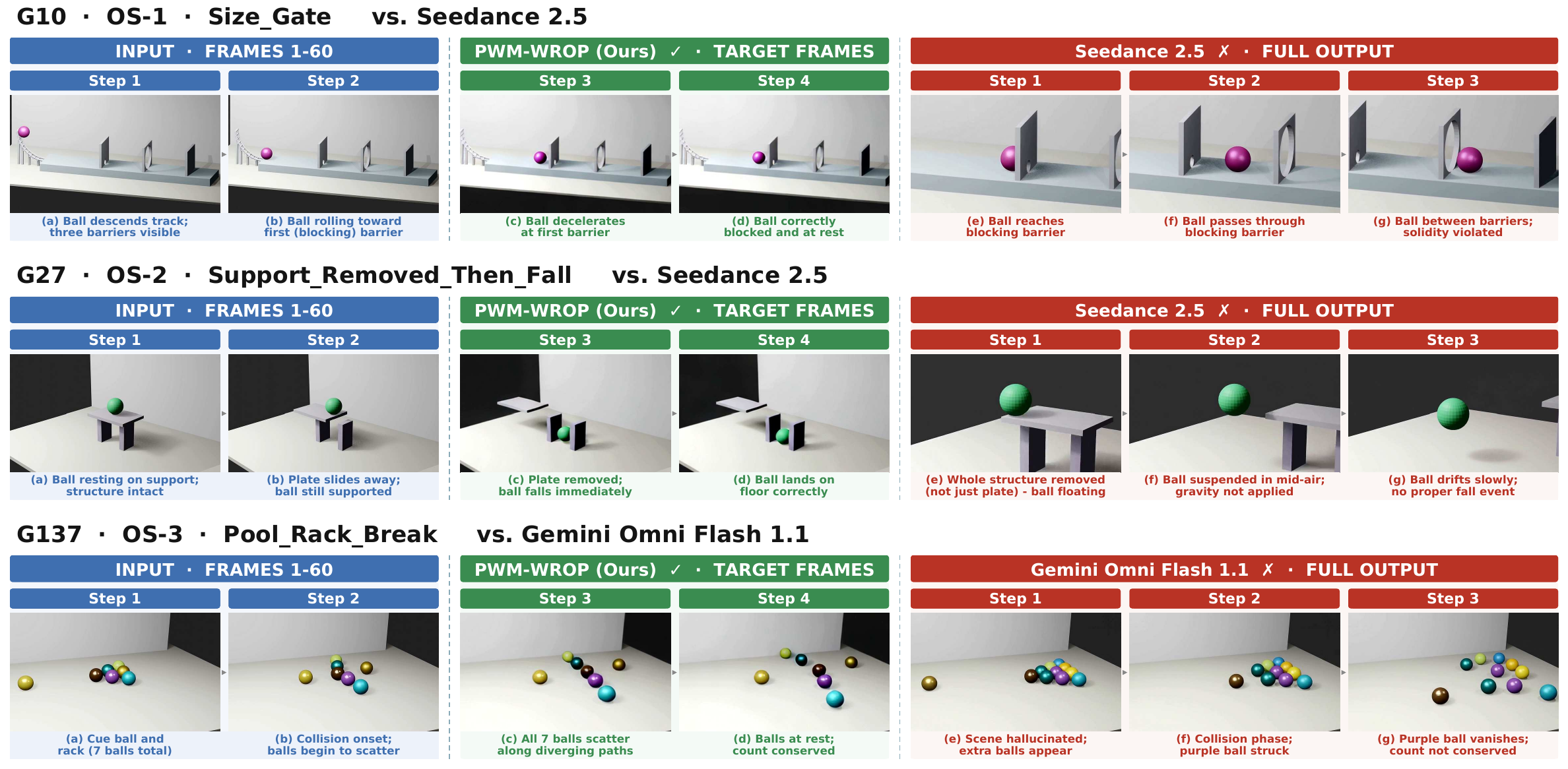}
\caption{Qualitative comparison on three object-solidity task families (Figure~\ref{fig:task-taxonomy}, bottom row). Layout follows Figure~\ref{fig:qualitative-op}: shared input (blue), PWM-WROP target output (green), and full baseline output (red). \textbf{G10 (OS-1):} Seedance 2.5 allows the ball to pass through the first barrier despite its aperture being half the ball diameter; PWM-WROP correctly brings the ball to rest at the barrier. \textbf{G27 (OS-2):} Seedance 2.5 removes the entire support structure rather than only the plate, then leaves the ball suspended in mid-air with no gravitational response; PWM-WROP generates an immediate, correctly-timed fall to the floor. \textbf{G137 (OS-3):} Gemini Omni Flash halluccinates additional balls before the collision, then allows the purple ball to vanish on impact, violating count conservation; PWM-WROP preserves all seven balls with physically plausible diverging trajectories.}
\label{fig:qualitative-os}
\end{figure*}

\textbf{Size-selective barrier passage (G10, OS-1).} \texttt{Size\_Gate} tests whether a model follows the physical principle that a solid object cannot pass through an aperture smaller than itself. The task places three barriers in sequence: the first has a hole half the ball's diameter (blocking), the second has a hole twice the ball's diameter (permitting), and the third is solid. A physically correct continuation must bring the ball to rest at the first barrier. PWM-WROP correctly decelerates the ball and generates a rest state against the blocking panel, showing that it follows the size relationship between the ball and the aperture. Seedance 2.5, by contrast, shows the most common failure observed across evaluated models--including MiniMax H3 and Wan 3.0 Prime--in which the ball passes through the blocking barrier as if the constraint were absent. The models generate the physically wrong trajectory because it does not account for whether the object's size is compatible with the aperture it encounters, treating the barrier as a visual element rather than a physical constraint.

\textbf{Support-contingent fall (G27, OS-2).} \texttt{Support\_Removed\_Then\_Fall} tests whether the ball falls when its support is removed. The input shows a ball resting on a plate; the plate slides away at the split boundary, and the target must show the ball falling immediately and landing on the floor. PWM-WROP correctly links the ball to the removal of the support: the ball begins to fall as the plate moves away and comes to rest on the floor in a single continuous motion. However, Seedance 2.5 shows two related failures. First, it removes the entire support structure rather than only the sliding plate, indicating that the model does not distinguish the plate from the legs and frame beneath it. Second, and more importantly, the ball remains suspended at its original height for several frames after the support structure has disappeared, then slowly begins to fall. This violates the solidity principle: once the support is removed, the ball should fall immediately. This suggests that the model does not correctly connect the removal of the support with the ball's subsequent fall.

\textbf{Count conservation under collision (G137, OS-3).} \texttt{Pool\_Rack\_Break} requires a model to generate a collision in which a rolling cue ball strikes a stationary rack of six balls, and all seven balls scatter outward along different paths, while preserving their identities and count. PWM-WROP generates a physically plausible break shot: all seven balls scatter in approximately correct directions, decelerate, and come to rest without any ball being created or disappearing. Gemini Omni Flash fails in two distinct ways. Before the collision, it generates additional balls in the rack, increasing the total count from seven to twelve or more. These extra balls are already visible in the first frames of its output, suggesting that the model does not keep track of the number of objects in the input scene. After the collision, the purple ball--the front ball of the rack and the first ball struck by the incoming cue ball--disappears from the scene. A ball disappearing on contact violates the solidity principle: solid objects should not disappear or merge when they collide; instead, they should transfer momentum and continue along separate paths. The combination of adding new balls and making an existing ball disappear indicates that Gemini Omni Flash does not consistently preserve the objects in the scene during a collision.

Taken together, these six cases reveal two types of failure that appear across both OP and OS tasks. The first is \emph{representation dropout}: the model generates a plausible-looking scene but fails to keep track of the objects that were present earlier. Examples include balls that do not re-emerge in their correct lanes, objects that suddenly appear after the occluder is removed, a ball revealed under the wrong cup, and a ball passing through a barrier that should block its path. The second is \emph{causal decoupling}: the model generates individual events that look reasonable but fails to connect them through the physical relationships between them. Examples include a ball that does not fall when its support is removed and a collision in which the number of balls changes after the impact. These two types of failure are also reflected in the family-level Elo results: Some models perform perform well on occlusion tracking, where the main challenge is to keep track of objects when they are hidden, but perform poorly on obstruction or collision tasks, where the model must account for how physical events affect subsequent motion, and vice versa. Closing the gap with human physical reasoning will require models to keep track of objects throughout the entire generated sequence, including what exists, where it is, and how physical interactions change their motion.

\subsection{Automatic Metrics}
\label{sec:res-auto}

\begin{table}[h]
\centering
\caption{Automatic full-reference metrics against the reference continuation, averaged over the 300 exam questions and ordered by human Elo (\cref{tab:elo}). For each model, the predicted span is extracted using its per-model trim rule, uniformly resampled to the 60 target frames, and compared at 320$\times$192; CLIP (ViT-B/32, 224$\times$224) and FID (299$\times$299, all frames at stride 2, $\approx$9{,}030 frames per side) are computed at their respective input resolutions. The best result in each column is shown in bold and the second-best is underlined. These metrics measure similarity to the target rather than object-permanence capability: edit/transfer models can repaint the static scene and score well on structural metrics without depicting the required reappearance.}
\label{tab:auto}
\small
\setlength{\tabcolsep}{4pt}
\begin{tabular}{lrrrrrrr}
\toprule
Model & LPIPS$\downarrow$ & MS-SSIM$\uparrow$ & SSIM$\uparrow$ & PSNR$\uparrow$ & MSE ($\times 10^{-3}$)$\downarrow$ & CLIP$\uparrow$ & FID$\downarrow$ \\
\midrule
Wan 3.0 Prime & 0.115 & 0.861 & \textbf{0.942} & \underline{27.15} & 4.48 & 0.948 & 15.1 \\
MiniMax H3 & \underline{0.105} & \underline{0.877} & \underline{0.938} & \textbf{27.51} & 9.09 & \textbf{0.962} & \underline{14.8} \\
\textbf{PWM-WROP (ours)} & \textbf{0.081} & \textbf{0.921} & 0.917 & 26.45 & \textbf{2.97} & \underline{0.956} & 20.4 \\
Seedance 2.5 & 0.282 & 0.616 & 0.770 & 18.37 & 28.55 & 0.928 & 24.6 \\
Runway Aleph 2\textsuperscript{\dag} & 0.181 & 0.789 & 0.918 & 24.98 & 9.71 & 0.918 & 19.7 \\
Wan-VACE 14B & 0.349 & 0.744 & 0.842 & 17.88 & 24.21 & 0.872 & 33.0 \\
Gemini Omni Flash 1.1 & 0.151 & 0.830 & 0.937 & 26.81 & \underline{3.88} & 0.943 & 17.0 \\
Kling O3 Pro & 0.163 & 0.818 & 0.933 & 26.10 & 4.73 & 0.932 & 18.0 \\
Grok Imagine (video extend) & 0.125 & 0.852 & 0.932 & 26.98 & 5.52 & 0.956 & \textbf{13.6} \\
LTX-2.3 Extend & 0.183 & 0.789 & 0.886 & 23.14 & 11.08 & 0.931 & 21.5 \\
Cosmos3 Super & 0.340 & 0.594 & 0.768 & 15.03 & 40.59 & 0.853 & 46.3 \\
LTX-2.3 Dev & 0.289 & 0.726 & 0.824 & 21.09 & 10.06 & 0.869 & 73.7 \\
HY-OmniWeaving & 0.247 & 0.716 & 0.874 & 22.41 & 17.61 & 0.875 & 43.8 \\
MAGI-1 24B & 0.260 & 0.681 & 0.853 & 21.59 & 20.66 & 0.923 & 26.6 \\
\bottomrule
\multicolumn{8}{l}{\footnotesize\textsuperscript{\dag}\,$n=297$: the Runway API refused three prompts longer than 1{,}000 characters; all other models $n=300$.} \\
\end{tabular}
\end{table}

Table~\ref{tab:auto} reports full-reference metrics against the target video, computed at $320{\times}192$ on each model's trimmed predicted span (Section~\ref{sec:protocol}). At this resolution, PWM-WROP performs best on perceptual distance (LPIPS 0.081; next best 0.105), structural similarity (MS-SSIM 0.921; next best 0.877), pixel error (MSE, MAE), and final-frame MSE, and ranks second on CLIP similarity. It ranks lower on PSNR and SSIM, and lower still on the resolution-sensitive measures computed at $720$p---final-frame LPIPS (0.166) and FID (20.444)---where its $320{\times}192$ output is upsampled four times before comparison.

These metrics measure how closely a generated clip matches a reference video rather than directly measuring object permanence or solidity reasoning, and differences between interface classes make them less suitable for ranking models. An edit model that reproduces the pre-event scene (reproducing its texture, lighting, and static geometry) may score well on SSIM and LPIPS without correctly generating the event required by the task. Conversely, a true-continuation model that correctly tracks an occluded object as it re-emerges may still differ from the reference in pixel space if its trajectory, especially its timing, differs from the ground truth. Full-reference metrics are therefore most useful as a diagnostic of visual and geometric consistency. For instance, they can confirm that PWM-WROP's low-resolution output does not lose absolute pixel fidelity compared with higher-resolution baselines when evaluated at the same scale, rather than serving as a primary measure of physical reasoning quality.
\section{Conclusion}

We introduced WROP, a benchmark and training resource grounded in the core-cognition framework, designed to probe whether video generation models have internalized the core representational constraints, object permanence and object solidity, that developmental science established to be foundational to physical intelligence. Built on 150 hand-designed Blender generators across six cognitively grounded task families, WROP provides a 1.5-million-sample training corpus and a fixed 300-question exam with human Elo ratings across 14 models. Fine-tuning PWM-WROP on this corpus yields the highest-ranked true-continuation model in a blind pairwise study with 20 raters, competitive with frontier commercial systems despite operating at a lower native resolution. The result offers preliminary evidence that training on cognitively principled synthetic data is a viable path toward enabling physical reasoning in video generation models. We release the corpus, exam, model answers, scores, weights, and PWM, our native-PyTorch training stack on AWS Trainium2, in the hope that WROP serves as a foundation for the community to measure, understand, and improve physical reasoning in video generation models.

\section*{Acknowledgements}

We thank Amazon Web Services for supporting this work through the AWS Trainium for Research program (\url{https://aws.amazon.com/ai/machine-learning/trainium/research/}).

\clearpage


\bibliography{main}
\bibliographystyle{icml2026}

\clearpage
\appendix
\section*{Appendix}

\section{Task Inventory}
\label{app:tasks}

\begingroup\scriptsize\setlength{\tabcolsep}{4pt}
\begin{longtable}{@{}llp{0.58\textwidth}@{}}
\caption{Task inventory of all 150 WROP generators across the six task families. Each entry names the generator and its core physical scenario.}\label{tab:task-inventory}\\
\toprule ID & Name & Description \\ \midrule \endfirsthead
\multicolumn{3}{@{}l}{\textit{Table~\ref{tab:task-inventory} continued}}\\ \toprule ID & Name & Description \\ \midrule \endhead
\midrule \multicolumn{3}{r@{}}{\textit{continued on next page}} \\ \endfoot
\bottomrule \endlastfoot
\multicolumn{3}{@{}l}{\textbf{OP-1: \texttt{Baillargeonian\_Occlusion}} (26 tasks)}\\*[1pt]
G02 & \texttt{ramp\_tunnel} & An object rolls down a ramp, passes through an opaque tunnel, reappears\,\ldots \\
G04 & \texttt{high\_low\_cover} & Objects pass behind high/low covers on crossing or straight paths; identity is preserved. \\
G14 & \texttt{u\_tube\_three\_lanes} & Three coloured balls travel parallel U-tube lanes, preserve identity\,\ldots \\
G17 & \texttt{moving\_tray} & A moving tray carries objects behind a fixed screen (one or two lanes). \\
G28 & \texttt{open\_ended\_tunnel} & An object travels through an open-ended tunnel, hidden in the middle. \\
G29 & \texttt{container\_entry\_left\_u} & One or two variably sized balls roll on two parallel left-U tracks. \\
G31 & \texttt{two\_lane\_tunnel} & Objects travel two-lane tunnels; identity is preserved. \\
G33 & \texttt{partial\_window\_parallel\_cars} & Parallel wheeled cars pass behind a partial window. \\
G35 & \texttt{single\_low\_window} & An object passes behind a low window. \\
G36 & \texttt{rollercoaster\_u\_track\_glassbox} & An object rolls a continuous U-track into a glass box. \\
G37 & \texttt{u\_track\_occluded\_ball} & A ball travels a U-track behind an occluder. \\
G38 & \texttt{pendulum\_occluded\_by\_screen} & A swinging pendulum is periodically occluded by a screen. \\
G39 & \texttt{lidded\_box\_ball\_enters} & A ball enters a lidded box and is hidden. \\
G42 & \texttt{ring\_track\_behind\_center\_block} & A ball on a ring track passes behind a center block. \\
G43 & \texttt{three\_balls\_parallel\_tunnels} & Three balls travel parallel tunnels; identity is preserved. \\
G54 & \texttt{serpentine\_ramp\_tunnel} & A ball rolls down a blue S-shaped serpentine ramp with three switchback segments\,\ldots \\
G61 & \texttt{spiral\_ramp\_behind\_column} & A ball rolls down a helical ramp that spirals around an opaque vertical column. \\
G62 & \texttt{two\_balls\_cross\_tunnel} & A red ball on the left and a blue ball on the right roll toward each other and both\,\ldots \\
G75 & \texttt{ball\_behind\_rotating\_billboard} & A ball rolls from left to right at a steady speed while remaining in contact with the\,\ldots \\
G76 & \texttt{pendulum\_behind\_post} & An opaque post stands in front of the lowest point of its arc\,\ldots \\
G77 & \texttt{three\_balls\_one\_wide\_tunnel} & Three balls --- red, green, blue, in that order --- roll one behind another along a\,\ldots \\
G78 & \texttt{rolling\_disc\_tunnel} & An upright disc (a coin standing on its edge) rolls along a straight flat track and\,\ldots \\
G101 & \texttt{picket\_fence\_flicker} & A ball rolls at a steady speed along a straight rail behind a row of evenly spaced\,\ldots \\
G102 & \texttt{drop\_screen\_occluder} & A ball rolls at a constant speed along a straight rail from left to right. \\
G103 & \texttt{corner\_turn\_occlusion} & A ball rolls along an L-shaped track. \\
G122 & \texttt{ball\_behind\_box\_stack} & A ball rolls horizontally across the table at a steady speed and passes behind a tall\,\ldots \\
\addlinespace[3pt]
\multicolumn{3}{@{}l}{\textbf{OP-2: \texttt{Object\_Static\_Occlusion}} (29 tasks)}\\*[1pt]
G15 & \texttt{ramp\_panel\_occludes\_objects} & Objects roll down a ramp behind a screen or panel and reappear. \\
G16 & \texttt{window\_slit\_mask\_reveals} & A window or slit mask scans across the scene, revealing objects piece by piece. \\
G18 & \texttt{turntable\_behind\_screen} & An object on a rotating turntable is partially occluded by a fixed screen and must\,\ldots \\
G19 & \texttt{row\_screen\_hides\_objects} & A sliding foreground screen hides a row of objects. \\
G20 & \texttt{connected\_vertical\_panel} & A connected vertical guide panel hides objects behind it. \\
G21 & \texttt{connected\_sliding\_doors} & Connected sliding doors close over a center object and reopen. \\
G22 & \texttt{top\_rail\_screen} & A top-rail-mounted screen slides across to hide objects. \\
G23 & \texttt{cabinet} & A display cabinet's cover or doors hide the objects inside. \\
G24 & \texttt{pivoting\_occluder} & A pivoting sign or hinged page swings shut to hide objects. \\
G25 & \texttt{no\_top\_side\_post\_panel} & A side-post-mounted panel hides objects behind it. \\
G26 & \texttt{bottom\_rail\_screen} & A bottom-rail-mounted screen slides across to hide objects. \\
G47 & \texttt{theater\_curtain} & A theater curtain closes over stage objects. \\
G48 & \texttt{sliding\_window\_row} & A sliding window reveals a row of objects. \\
G49 & \texttt{two\_supported\_screens\_close} & Two supported screens close and reopen over objects. \\
G51 & \texttt{vertical\_panel\_two\_objects} & A vertical rail panel hides two objects. \\
G53 & \texttt{no\_gap\_top\_rail\_curtain} & A no-gap top-rail curtain hides three balls. \\
G55 & \texttt{rising\_floor\_screen} & Two colored blocks rest side by side on a table. \\
G63 & \texttt{comb\_occluder\_sweep} & Three coloured blocks sit in a row on a table. \\
G64 & \texttt{flipboard\_occluder} & A flat board, hinged along its bottom edge in front of them\,\ldots \\
G79 & \texttt{venetian\_blinds} & The slats all rotate to lie flat and fully occlude the objects, hold\,\ldots \\
G80 & \texttt{sliding\_double\_doors} & Two opaque doors slide in from the left and right until they meet and fully cover the\,\ldots \\
G81 & \texttt{rolling\_shutter} & A segmented rolling shutter descends from above in front of them\,\ldots \\
G82 & \texttt{accordion\_fold\_screen} & A folded accordion screen at one side unfolds sideways across in front of them until\,\ldots \\
G104 & \texttt{bifold\_concertina\_doors} & A pair of bi-fold doors, hinged in the middle, unfold from the left and right until\,\ldots \\
G105 & \texttt{descending\_dome\_cover} & A dome-shaped cover lowers straight down from above to fully enclose and hide them\,\ldots \\
G106 & \texttt{rotating\_drum\_occluder} & The drum rotates about its vertical axis until its solid wall faces the camera\,\ldots \\
G121 & \texttt{wiper\_screen\_occlusion} & A tall vertical screen, hinged at its base on one side\,\ldots \\
G123 & \texttt{sliding\_cover\_panel} & A flat upright cover panel standing on a low track slides sideways in front of the\,\ldots \\
G124 & \texttt{rising\_sleeve\_cover} & An open cylindrical sleeve, visibly wider than the pedestal\,\ldots \\
\addlinespace[3pt]
\multicolumn{3}{@{}l}{\textbf{OP-3: \texttt{Container\_Permanence}} (35 tasks)}\\*[1pt]
G08 & \texttt{drawer\_moves\_hidden\_object} & A closing drawer carries a hidden object as it slides. \\
G11 & \texttt{box\_holds\_N\_objects} & A lidded box hides one to three objects through a rotation. \\
G12 & \texttt{partition\_box} & A partitioned box hides several objects in a fixed order. \\
G13 & \texttt{three\_drawers} & Three drawers hide objects in a fixed order. \\
G30 & \texttt{hidden\_object\_moves\_with\_cart} & A moving cart carries a hidden object. \\
G32 & \texttt{cart\_swap} & Two carts swap positions while carrying hidden objects. \\
G40 & \texttt{guided\_elevator\_hidden\_ball} & A guided elevator carries a hidden ball upward. \\
G45 & \texttt{opaque\_lid\_front\_panel\_drops} & An opaque lid or front panel drops over a center ball. \\
G46 & \texttt{marked\_boxes\_swap} & Marked boxes swap positions while hiding their objects. \\
G56 & \texttt{three\_cup\_shell\_game} & Three identical opaque cups upside-down on a table\,\ldots \\
G59 & \texttt{hinged\_lid\_box\_opens} & A closed gray box whose top lid, hinged at the back edge\,\ldots \\
G65 & \texttt{box\_two\_balls\_relocate} & A lid closes over the box, hiding the balls; the box then slides across the table to a\,\ldots \\
G66 & \texttt{rotating\_carousel\_cups} & The turntable rotates 180 degrees, carrying all three cups around with it. \\
G71 & \texttt{double\_doors\_swing\_open} & The two doors swing open outward on their side hinges --- the left door to the left\,\ldots \\
G72 & \texttt{sliding\_lid\_box} & Its flat top lid slides off horizontally to one side\,\ldots \\
G83 & \texttt{four\_cup\_three\_swaps} & The cups slide through three sequential position swaps. \\
G84 & \texttt{two\_balls\_three\_cups} & The cups slide through two position swaps. \\
G85 & \texttt{nested\_cup\_transfer} & A ball is shown, then a small opaque cup lowers over it. \\
G86 & \texttt{conveyor\_covered\_boxes} & Four identical opaque covers sit in a row on a conveyor belt\,\ldots \\
G87 & \texttt{four\_cup\_carousel} & The turntable rotates 180 degrees, carrying all four cups around. \\
G88 & \texttt{shell\_game\_fakeout\_reveal} & The cups slide through two position swaps. \\
G96 & \texttt{clamshell\_box} & Its top half and bottom-front half swing open about a rear hinge --- the top lifting up\,\ldots \\
G97 & \texttt{rolltop\_tambour} & A box with a curved roll-top (tambour) cover sits on a table. \\
G98 & \texttt{liftoff\_dome\_lid} & The dome lifts straight up, clearing the object, to reveal it resting on the base. \\
G107 & \texttt{sliding\_cup\_relocate} & A ball rests on a table; a single opaque cup is lowered over it\,\ldots \\
G108 & \texttt{two\_carts\_cross\_swap} & Two identical covered carts sit at opposite ends of a table\,\ldots \\
G109 & \texttt{tilting\_tray\_relocate} & A lid closes over it; the tray then tilts so the hidden object slides under the cover\,\ldots \\
G116 & \texttt{vault\_swing\_door} & A heavy round vault door on a side hinge is closed over the front of a safe. \\
G117 & \texttt{blooming\_petal\_box} & Four triangular flaps are folded up and inward to form a closed pyramid over an object. \\
G118 & \texttt{matchbox\_drawer} & The inner tray then slides straight back INTO the sleeve horizontally\,\ldots \\
G125 & \texttt{two\_carts\_reveal\_empty} & Two identical covered carts sit at opposite ends of a table. \\
G126 & \texttt{turntable\_two\_boxes} & The front box's cover lifts to show a ball under it, then lowers to hide it. \\
G131 & \texttt{stacked\_drawers\_reveal} & The lower drawer slides out toward the viewer, carrying the ball\,\ldots \\
G132 & \texttt{twist\_open\_capsule} & The top half lifts straight up and off with a slight twist to reveal the ball resting\,\ldots \\
G144 & \texttt{double\_flap\_top\_box} & Starting open, both flaps swing down together and close over the ball\,\ldots \\
\addlinespace[3pt]
\multicolumn{3}{@{}l}{\textbf{OS-1: \texttt{Baillargeonian\_Obstruction}} (20 tasks)}\\*[1pt]
G03 & \texttt{ramp\_ball\_blocked\_by\_wall} & An object rolls down a ramp into a solid or transparent wall and is blocked without\,\ldots \\
G05 & \texttt{car\_vs\_barrier} & A vehicle meets a wall, hole, gate, or door --- passing through an opening or being blocked. \\
G07 & \texttt{guillotine\_gate\_stops} & A vehicle or ball is stopped by a descending guillotine gate. \\
G09 & \texttt{low\_beam\_car\_height} & A tall vehicle is blocked by a low beam while a short one passes underneath. \\
G10 & \texttt{size\_gate} & A small object passes through an aperture; a larger one is blocked, rebounds slightly\,\ldots \\
G34 & \texttt{occ\_small\_ball\_behind\_screen} & A small ball passes behind a foreground screen. \\
G50 & \texttt{ramp\_ball\_slotted\_screen} & A ramp ball passes a slotted screen. \\
G52 & \texttt{ramp\_open\_right\_gated\_box} & A ball rolls down a right-opening ramp into a gated catch-box. \\
G57 & \texttt{ramp\_ball\_deflected\_angled\_wall} & A ball rolls down a ramp toward a fixed gray wall angled at 45 degrees and deflects\,\ldots \\
G67 & \texttt{double\_deflector\_zigzag} & A ball is released and falls onto a left-leaning angled wall\,\ldots \\
G68 & \texttt{width\_slot\_wall\_two\_balls} & A standing wall blocks a flat track, but it has a narrow vertical slot cut through its\,\ldots \\
G89 & \texttt{limbo\_height\_bar} & A short object rolls along the track and passes cleanly UNDER the bar to the far side\,\ldots \\
G90 & \texttt{funnel\_size\_sorter} & A small ball dropped into the funnel rolls down to the apex and passes through the\,\ldots \\
G91 & \texttt{turnstile\_timed\_gate} & A ball rolls toward it and arrives when a gap between two arms is aligned with the\,\ldots \\
G92 & \texttt{bumper\_carom} & A ball rolls across a flat surface toward a fixed round bumper post (a cylinder). \\
G110 & \texttt{one\_way\_flap\_gate} & A ball rolling in from the open side pushes the flap open and passes through\,\ldots \\
G111 & \texttt{portcullis\_drop\_gate} & One ball rolls under it while it is up and passes through. \\
G112 & \texttt{banked\_quarter\_pipe\_redirect} & A ball rolls toward a curved banked wall (a quarter-pipe). \\
G127 & \texttt{rising\_bollard\_stop} & A ball rolls along the table toward a spot at constant speed. \\
G128 & \texttt{swing\_arm\_barrier\_stop} & A ball rolls up to the closed arm and is stopped -- it decelerates to rest against the\,\ldots \\
\addlinespace[3pt]
\multicolumn{3}{@{}l}{\textbf{OS-2: \texttt{Object\_Drop}} (21 tasks)}\\*[1pt]
G01 & \texttt{hole\_box\_drop} & An object drops onto a box with a covered or open center hole --- it rests on the cap or\,\ldots \\
G06 & \texttt{car\_on\_bridge} & A vehicle crosses a bridge that has a hole or a solid span; it falls through or passes. \\
G27 & \texttt{support\_removed\_then\_fall} & A support is removed and the object falls. \\
G41 & \texttt{trapdoor\_opens\_ball\_falls} & A trapdoor opens, the ball falls through to the floor\,\ldots \\
G58 & \texttt{two\_ball\_trapdoor\_size\_filter} & A small ball and a large ball roll onto a gray platform with a circular hole in the\,\ldots \\
G69 & \texttt{two\_tier\_trapdoor\_cascade} & The trapdoor opens and the ball falls through onto a lower platform below\,\ldots \\
G70 & \texttt{popaway\_support\_columns} & The two columns slide out sideways from under the slab\,\ldots \\
G93 & \texttt{tipping\_shelf\_drop} & The prop slides out; with that end unsupported the shelf tips down\,\ldots \\
G94 & \texttt{trapdoor\_drop} & The trapdoor swings open downward on its hinge, removing the support\,\ldots \\
G95 & \texttt{conveyor\_edge\_fall} & When it reaches the edge, it rolls off and falls into a bin (a lower catch floor) below. \\
G113 & \texttt{whipped\_away\_card} & The card is flicked out sideways fast; the ball, left unsupported\,\ldots \\
G114 & \texttt{retractable\_support\_pins} & The two pins retract sideways into the wall, removing all support\,\ldots \\
G115 & \texttt{steep\_slope\_slide} & It slides and rolls down the slope, accelerating under gravity, reaches the bottom\,\ldots \\
G129 & \texttt{sliding\_hatch\_drop} & The hatch slides sideways fully clear of the hole, removing the support\,\ldots \\
G130 & \texttt{bomb\_bay\_doors\_drop} & The two doors swing downward and apart on their outer hinges (like bomb-bay doors)\,\ldots \\
G145 & \texttt{tilt\_platform\_rolloff} & The platform tilts down on one side; the ball rolls to the low edge, rolls off\,\ldots \\
G146 & \texttt{hanging\_ball\_release\_drop} & The hook releases and the string goes slack, so the ball drops straight down to the\,\ldots \\
G147 & \texttt{drop\_leaf\_shelf} & A ball rests on a shelf hinged at the back wall, held out horizontally like a\,\ldots \\
G148 & \texttt{latch\_release\_flap\_drop} & The pin is pulled out sideways; released, the flap flips down about its hinge and the\,\ldots \\
G149 & \texttt{snap\_pillar\_topple} & Losing its support, the ball drops nearly straight down to the floor and bounces to rest. \\
G150 & \texttt{rollers\_part\_drop} & A ball rests nestled in the valley between two parallel horizontal rollers. \\
\addlinespace[3pt]
\multicolumn{3}{@{}l}{\textbf{OS-3: \texttt{Object\_Collision}} (19 tasks)}\\*[1pt]
G44 & \texttt{two\_balls\_collide\_and\_bounce} & Two balls collide and bounce apart. \\
G60 & \texttt{newtons\_cradle} & The leftmost ball is raised and released; it strikes the row and the rightmost ball\,\ldots \\
G73 & \texttt{break\_scatter\_cluster} & A cue ball rolls across a flat surface into a tight triangular cluster of stationary\,\ldots \\
G74 & \texttt{glancing\_oblique\_collision} & A moving ball rolls across a flat surface and strikes a stationary ball off-centre (a\,\ldots \\
G99 & \texttt{headon\_velocity\_exchange} & A ball rolls straight along a line into a second, identical stationary ball. \\
G100 & \texttt{heavy\_light\_collision} & A large heavy ball rolls into a small light stationary ball. \\
G119 & \texttt{offcenter\_break\_vsplit} & A ball rolls into the seam of two touching balls at rest, striking them off-center. \\
G120 & \texttt{pendulum\_strike\_projectile} & A heavy metal ball hangs as a pendulum, is raised and released\,\ldots \\
G133 & \texttt{ball\_topples\_block} & A ball rolls across the flat floor and strikes a single standing block. \\
G134 & \texttt{knock\_ball\_off\_tee} & A second ball rolls across the flat surface and strikes the resting ball. \\
G135 & \texttt{lever\_launch\_transfer} & A ball drops onto the raised end of a see-saw lever resting on a fulcrum. \\
G136 & \texttt{ball\_strikes\_pendulum} & A second ball rolls in along the table and strikes the hanging ball at the bottom of\,\ldots \\
G137 & \texttt{pool\_rack\_break} & A ball rolls across a flat surface and strikes a triangular rack of six resting balls. \\
G138 & \texttt{bank\_shot\_carom} & A ball rolls across a flat surface, banks off a straight cushion at an angle (a clean\,\ldots \\
G139 & \texttt{ball\_shoves\_block\_slide} & A ball rolls across a flat surface toward a block resting upright on the table. \\
G140 & \texttt{light\_ball\_rebounds\_off\_heavy} & A small light ball rolls into a large heavy stationary ball. \\
G141 & \texttt{glancing\_billiard\_split} & A ball rolls across a flat surface and strikes a resting ball off-centre. \\
G142 & \texttt{knock\_ball\_off\_ledge} & A ball rolls along the top of a raised platform toward a second ball resting at the\,\ldots \\
G143 & \texttt{topple\_two\_blocks\_apart} & Two tall blocks stand upright side by side on the table with a small gap between them. \\
\end{longtable}
\endgroup

\clearpage

\section{PWM on Trainium2}
\label{app:infra}

\paragraph{Overview.} The released \texttt{pwm} package trains and samples Cosmos3-Nano on AWS Trainium2 in native PyTorch without XLA graph tracing. It loads the public diffusers-layout weights directly, shards them across a two-dimensional device mesh of tensor parallelism $\times$ FSDP2 using DTensor, and compiles each Mixture-of-Transformers block once with static shapes. Training uses an fp32 master copy under FSDP2 with bf16 compute; checkpoints are written per rank, support bit-exact resumption, and can be consolidated back to the diffusers layout. A single YAML file specifies the geometry and parallel layout; the CLI re-launches itself under \texttt{torchrun} with $\mathrm{tp}\times\mathrm{fsdp}$ processes. All detectable misconfigurations are rejected before the first training step, including: non-integer patch shapes; a tensor-parallel degree that does not divide the head count; an empty checkpoint directory; bf16 training without an fp32 master; a process count inconsistent with the mesh; clips whose geometry does not match the declared configuration; a resumed run whose configuration differs from the checkpoint; and a checkpoint directory too small for the run.

\paragraph{Hardware.} Experiments run on a single \texttt{trn2.48xlarge}: sixteen Trainium2 chips exposed as 64 logical NeuronCores connected by NeuronLink. The production layout for the 36-layer model uses tensor parallelism degree 4 and FSDP degree 16 over the 64 cores. Inference runs on the tensor-parallel group alone (bf16, 29\,GB of weights, 7.3\,GB per core) and therefore also fits a four-core \texttt{trn2.3xlarge}.

\paragraph{Throughput.} Table~\ref{tab:trainium} reports measured training step times at the 288$\times$512, 30-latent-frame geometry (117 video frames; 4,448 tokens per sample: 128 text and 4,320 vision; batch size 16). Over three successive optimisations---each verified to leave the loss trajectory unchanged step for step---the step time decreased from 15.1\,s to 5.7\,s. 

\begin{table}[h]
\centering
\small
\caption{Training step time on one \texttt{trn2.48xlarge} (36-layer Cosmos3-Nano, tp4$\times$fsdp16, batch~16, $288\times512\times30$ latent frames; medians over 10 steps after 3 warm-up steps). Loss trajectories are identical step for step across all rows.}
\label{tab:trainium}
\begin{tabular}{lrr}
\toprule
\textbf{Configuration} & \textbf{s/step} & \textbf{tokens/s} \\
\midrule
First working 64-core run & 15.1 & 4,705 \\
+ reduce-scatter copy-in rewritten (row concatenation) & 6.74 & 10,564 \\
+ multi-tensor AdamW, one sync per step; FSDP2 prefetch depth 2 & 5.71 & 12,454 \\
\bottomrule
\end{tabular}
\end{table}

\paragraph{Step time breakdown.} At 6.74\,s per step, the block forward and backward account for 3.4\,s; the remainder comprises FSDP2 all-gathers (1.8\,s over 75 calls), reduce-scatters (1.4\,s), the root unit's all-gather with embedding, head, and loss (1.25\,s), the optimiser (1.25\,s), and gradient clipping (0.37\,s), with some overlap between these stages. Reducing the FSDP degree does not improve throughput: the collectives are latency-bound (200\,MB in 24\,ms), so gains come from eliminating collectives rather than reducing their payload.

\paragraph{Engineering findings.} The following findings emerged during development and apply broadly to large models trained in native PyTorch on this backend.

\begin{itemize}[leftmargin=*,itemsep=2pt,topsep=2pt]
\item \textbf{Load-then-shard fragments device memory.} Materialising the full 14.6\,GB fp32 tensor-parallel shard before FSDP2 partitions it leaves insufficient memory for the root unit's 2.5\,GB reduce buffer, causing the first backward pass to fail. Modules are therefore loaded and sharded in an interleaved pass.

\item \textbf{FSDP2's reduce-scatter copy-in is the dominant bottleneck.} The default chunk-concatenation copy-in consumed 7.9\,s of a 15.9\,s step; replacing it with a row-concatenation variant---verified to produce identical outputs---reduced step time by $2.25\times$.

\item \textbf{Replicated gradients drift under tensor parallelism.} Parameters replicated across the tensor-parallel group diverged by $5.9\times10^{-5}$ relative after 100 steps. Their gradients are now explicitly synchronised at every step.

\item \textbf{Single-tensor AdamW is launch-bound.} The default implementation incurs approximately ten kernel launches and 22 host synchronisations per parameter. A multi-tensor implementation with one synchronisation per step reduces optimiser time from 1,324\,ms to 126\,ms per step on a rank holding 662 tensors, with parameters and moments bitwise-equal over five steps.

\item \textbf{The replicated vocabulary embedding blocks the next optimisation.} Keeping blocks unsharded between forward and backward fails on the root unit's 2.49\,GB fp32 reduce buffer for the 151,936$\times$4,096 embedding, which is replicated across the tensor-parallel group. The two available fixes are to shard the vocabulary across the group or to freeze the text embedding during fine-tuning; this remains an open item at the time of writing.

\item \textbf{Fused attention kernels have geometry constraints.} A Neuron Kernel Interface flash-attention path is implemented, but its backward pass is limited to sequences of at most 8,192 tokens, excluding 720p geometries. It has not yet been A/B-tested against the compiled scaled-dot-product path end to end; the released default is the latter.

\item \textbf{Compiled blocks must not fall back to host execution.} Each block is compiled once with static shapes; any operator that falls back to the host inflates a sub-second block to several seconds. Input prompts are padded to the training text length with padding masked from the vision tokens so that every input compiles to a single shape.
\end{itemize}

\paragraph{Correctness validation.} Before throughput measurement, the port passed the following gates: bitwise forward parity with the upstream reference on CPU and two-layer real-weight parity of $7\times10^{-7}$; a 64-rank bit-exact resume; overfitting on six upstream example clips (sample-to-source cosine 0.979) and on 64 clips from our physics renders under the video-to-video protocol (continuation-half cosine 0.993--0.999 against 0.85--0.995 for the base model, conditioning half locked at 1.000); and same-seed sampling agreement with an independent earlier port (latent cosine 0.88--0.97 on five prompts, with residual differences attributable to kernel-level numerical variation compounded over 35 sampler steps). A full 64-rank checkpoint of the 36-layer model occupies 162\,GiB.
\clearpage
\section{Detailed Per-family Human Preference Data and Leaderboards}
\label{app:family-elo}

\begin{figure*}[h]
\centering
\includegraphics[width=\textwidth]{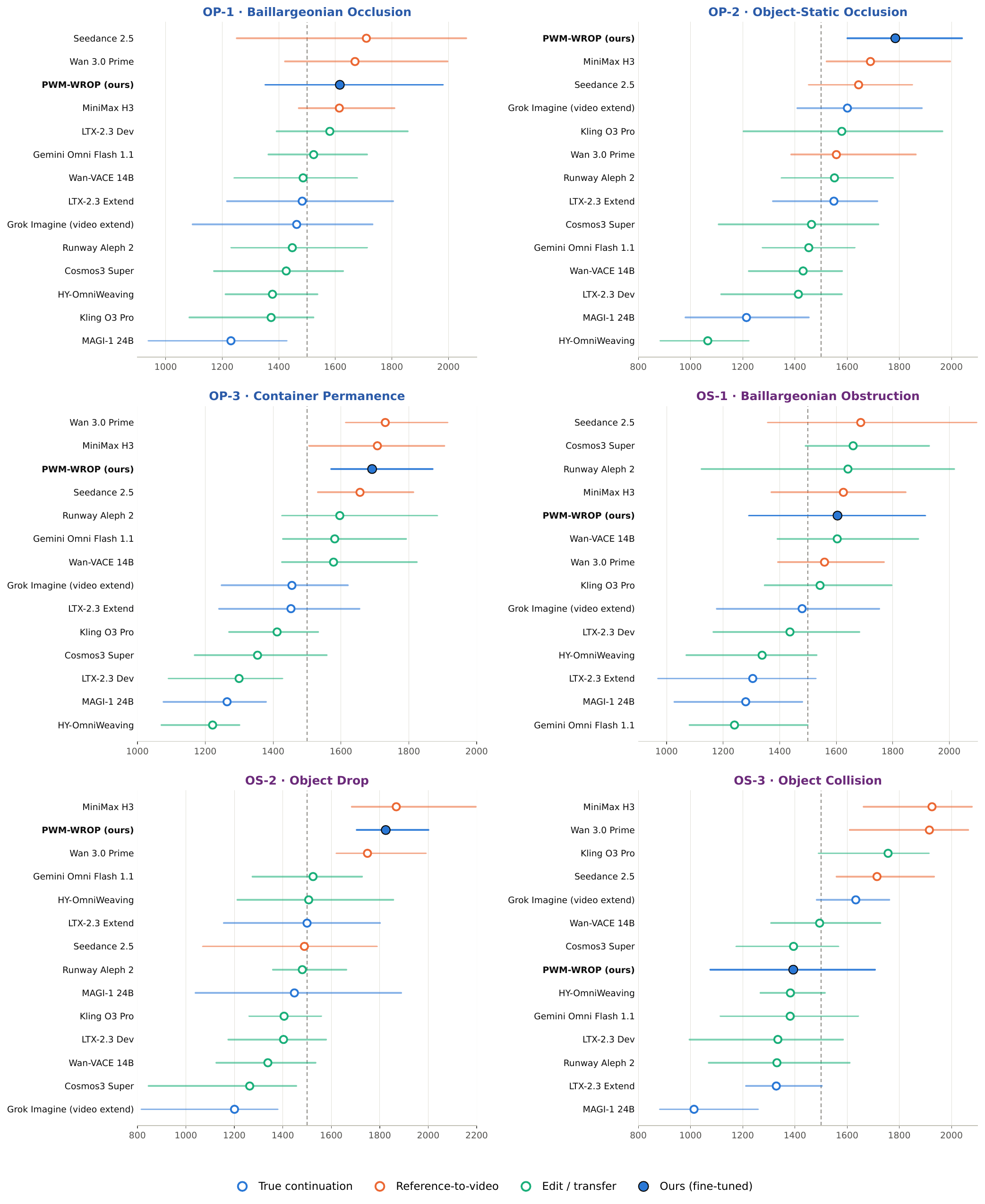}
\caption{Per-family human-preference Elo leaderboards for all six WROP task families. Points are Bradley--Terry MLE strengths on an Elo scale (mean 1500, dashed line); bars are 95\% rater-clustered bootstrap intervals. Colour encodes interface class as in Figure~\ref{fig:elo}: filled blue circle is PWM-WROP (ours); open circles are other true-continuation models (blue), reference-to-video models (orange), and edit/transfer models (green). Each panel is sorted independently by within-family rank. Intervals are especially wide in OS families (36--44 total games each); adjacent ranks are rarely distinguishable. Numerical values and within-family ranks are reported in Table~\ref{tab:family-elo}.}
\label{fig:family-leaderboards}
\end{figure*}

Table~\ref{tab:family-elo} reports Bradley--Terry Elo and within-family rank for all fourteen models across each of the six task families, and Figure~\ref{fig:family-leaderboards} plots the corresponding forest plots per family. Bootstrap intervals are correspondingly wide, and the patterns discussed in Section~\ref{sec:res-family} are most reliable where they are consistent across multiple families or corroborated by the qualitative analysis.

\begin{table}[h]
\centering
\footnotesize
\setlength{\tabcolsep}{5pt}
\renewcommand{\arraystretch}{1.08}
\caption{Per-family Bradley--Terry Elo for all 14 models across the six WROP task families. Superscripts give within-family rank (1~=~best). \underline{Underline} marks the top-ranked model in each family; \textbf{bold} marks PWM-WROP (ours). Overall Elo is from the joint fit over all 361 judgments; models are ordered by overall Elo rank. Bootstrap CIs are wide especially in OS families (36--44 total games each); ranks indicate tendency rather than significant differences.}
\label{tab:family-elo}
\resizebox{\linewidth}{!}{%
\begin{tabular}{@{}lcllllllll@{}}
\toprule
Model & Class & \multicolumn{1}{l}{Overall} & \multicolumn{3}{c}{Object Permanence} & \multicolumn{3}{c}{Object Solidity} \\
\cmidrule(lr){4-6}\cmidrule(lr){7-9}
 & & \multicolumn{1}{l}{Elo} & OP-1 & OP-2 & OP-3 & OS-1 & OS-2 & OS-3 \\
\midrule
Wan 3.0 Prime & Ref.-to-video & 1723.6 & 1670$^{2}$ & 1559$^{6}$ & \underline{1730}$^{1}$ & 1558$^{7}$ & 1749$^{3}$ & 1915$^{2}$ \\
MiniMax H3 & Ref.-to-video & 1723.6 & 1614$^{4}$ & 1689$^{2}$ & 1707$^{2}$ & 1625$^{4}$ & \underline{1868}$^{1}$ & \underline{1925}$^{1}$ \\
\addlinespace[4pt]
\textbf{PWM-WROP (ours)} & \textbf{True cont.} & \textbf{1679.5} & \textbf{1616}$^{3}$ & \textbf{1785}$^{1}$ & \textbf{1692}$^{3}$ & \textbf{1604}$^{5}$ & \textbf{1825}$^{2}$ & \textbf{1394}$^{8}$ \\
\addlinespace[4pt]
Seedance 2.5 & Ref.-to-video & 1649.6 & \underline{1710}$^{1}$ & 1644$^{3}$ & 1656$^{4}$ & \underline{1687}$^{1}$ & 1489$^{7}$ & 1714$^{4}$ \\
\addlinespace[4pt]
Runway Aleph 2 & Edit/transfer & 1518.3 & 1448$^{10}$ & 1552$^{7}$ & 1596$^{5}$ & 1641$^{3}$ & 1480$^{8}$ & 1331$^{12}$ \\
Wan-VACE 14B & Edit/transfer & 1506.7 & 1486$^{7}$ & 1431$^{11}$ & 1578$^{7}$ & 1604$^{6}$ & 1338$^{12}$ & 1495$^{6}$ \\
Gemini Omni Flash 1.1 & Edit/transfer & 1492.5 & 1523$^{6}$ & 1453$^{10}$ & 1582$^{6}$ & 1240$^{14}$ & 1525$^{4}$ & 1382$^{10}$ \\
Kling O3 Pro & Edit/transfer & 1471.3 & 1373$^{13}$ & 1580$^{5}$ & 1412$^{10}$ & 1543$^{8}$ & 1405$^{10}$ & 1757$^{3}$ \\
\addlinespace[4pt]
Grok Imagine (extend) & True cont. & 1457.0 & 1463$^{9}$ & 1601$^{4}$ & 1455$^{8}$ & 1479$^{9}$ & 1201$^{14}$ & 1633$^{5}$ \\
LTX-2.3 Extend & True cont. & 1453.4 & 1483$^{8}$ & 1549$^{8}$ & 1452$^{9}$ & 1305$^{12}$ & 1500$^{6}$ & 1329$^{13}$ \\
\addlinespace[4pt]
Cosmos3 Super & Edit/transfer & 1409.2 & 1426$^{11}$ & 1464$^{9}$ & 1354$^{11}$ & 1660$^{2}$ & 1263$^{13}$ & 1395$^{7}$ \\
LTX-2.3 Dev & Edit/transfer & 1398.9 & 1580$^{5}$ & 1413$^{12}$ & 1300$^{12}$ & 1436$^{10}$ & 1403$^{11}$ & 1335$^{11}$ \\
HY-OmniWeaving & Edit/transfer & 1268.5 & 1378$^{12}$ & 1066$^{14}$ & 1222$^{14}$ & 1338$^{11}$ & 1507$^{5}$ & 1383$^{9}$ \\
\addlinespace[4pt]
MAGI-1 24B & True cont. & 1248.0 & 1231$^{14}$ & 1214$^{13}$ & 1264$^{13}$ & 1280$^{13}$ & 1448$^{9}$ & 1013$^{14}$ \\
\bottomrule
\end{tabular}}
\end{table}

\end{document}